 \documentclass[preprint,3p,times]{elsarticle}

\usepackage[utf8]{inputenc} 
\usepackage[T1]{fontenc}    
\usepackage{hyperref}       
\usepackage{url}            
\usepackage{booktabs}       
\usepackage{amsfonts}       
\usepackage{nicefrac}       
\usepackage{microtype}      
\usepackage{graphicx}

\usepackage{indentfirst,csquotes}
\usepackage{tikz}
\usetikzlibrary{external}
\usepackage{siunitx}
\usepackage{graphicx}
\usetikzlibrary{positioning, fit, arrows.meta, shapes, decorations.pathreplacing}
\usetikzlibrary{spy}

\usepackage{amssymb,amsthm,amsmath}
\usepackage{xcolor,paralist,hyperref,titlesec,fancyhdr,etoolbox}

\journal{Computers and Fluids}

\begin{document}
\begin{frontmatter}
\title{Disentangled Latent Spaces for Reduced Order Models using Deterministic Autoencoders}
\author[inst1]{Henning Schwarz\corref{cor1}}
\affiliation[inst1]{Institute for Fluid Dynamics and Ship Theory, Hamburg University of Technology, Am Schwarzenberg-Campus 4, 21073 Hamburg, Germany}
\author[inst1]{Pyei Phyo Lin}
\author[inst2]{Jens-Peter M. Zemke}
\author[inst1]{Thomas Rung}
\affiliation[inst2]{Institute of Mathematics, Hamburg University of Technology, Am Schwarzenberg-Campus 3, 21073 Hamburg, Germany}

\cortext[cor1]{Corresponding author. \textit{E-mail address: }henning.schwarz@tuhh.de}


\begin{abstract}
Data-driven reduced-order models based on autoencoders generally lack interpretability compared to classical methods such as the proper orthogonal decomposition. More interpretability can be gained by disentangling the latent variables and analyzing  the resulting modes. For this purpose, probabilistic $\beta$-variational autoencoders ($\beta$-VAEs) are frequently used in computational fluid dynamics and other simulation sciences. 
Using a benchmark periodic flow dataset, we show that competitive results can be achieved using non-probabilistic autoencoder approaches that either promote orthogonality or penalize correlation between latent variables.
Compared to probabilistic autoencoders, these approaches offer more robustness with respect to the choice of hyperparameters entering the loss function. 
We further demonstrate the ability of a non-probabilistic approach to identify a reduced number of active latent variables by introducing a correlation penalty, a function also known from the use of $\beta$-VAE. The investigated probabilistic and non-probabilistic autoencoder models are finally used for the dimensionality reduction of aircraft ditching loads, which serves as an industrial application in this work.
\end{abstract} 


\end{frontmatter}

\section{Introduction}
Machine learning (ML) approaches based on either classical dimension reduction methods like the proper orthogonal decomposition (POD)~\cite{lumley1967structure} or autoencoders are frequently used to reduce the dimension of fluid flows in data-driven surrogate models~\cite{lusch:2018, Swischuk.2019, eivazi:2020, Agostini20, GANTI2020104626, wu:2021}.
An autoencoder passes the input through a dimension-reducing encoder and subsequently a dimension-increasing decoder. It aims to learn a lower-dimensional representation of the input by forcing the output of the network to be close to the input.  When considering two- or three-dimensional data, which is often the case in physics, autoencoders are usually used with convolutional layers to capture local dependencies in the data. 
To predict the temporal dynamics of unsteady physical phenomena, autoencoders can be combined with networks suitable for time series prediction like a long short-term memory (LSTM) network~\cite{hochreiter:1997}, a transformer~\cite{vaswani:2017} and other ML frameworks such as the sparse identification of nonlinear dynamics (SINDy)~\cite{brunton2016discovering} or approximations to the Koopman operator~\cite{Koopman:1931}. The dynamics are then usually learned in the obtained lower-dimensional latent space of the autoencoder~\cite{lusch:2018, eivazi:2020, wu:2021, GUPTA2022105239, Hemmasian:2023, ANDO2023106047, ZHANG2023105883, solera-rico:2024, schwarz:2024}.

By using nonlinear activation functions, an autoencoder performs nonlinear dimension reduction in contrast to the linear POD framework. Due to this, autoencoders usually achieve better reconstruction accuracy and a more compact latent space compared to POD. This comes, however, at the cost of losing interpretability, as the latent variables of the autoencoder are generally not disentangled or not orthogonal. 
The difficulty is that the influence of individual latent variables is hard to assess if the change in one latent variable inevitably forces a change in another latent variable due to entanglement.
Therefore, several studies have recently focused on obtaining disentangled (orthogonal) latent variables in autoencoders to gain interpretability~\cite{eivazi:2022, kang:2022, solera-rico:2024}. 
There are several strategies to achieve this goal.
The $\beta$-variational autoencoder ($\beta$-VAE)~\cite{Higgins2016betaVAELB}, which is based on the variational autoencoder (VAE)~\cite{Kingma:2013}, is probably the most used option. It is a probabilistic framework that aims at normally distributed latent space variables.  
A hyperparameter $\beta>0$ in the loss function controls the importance of reconstructive capability, typically in terms of the mean squared error, relative to a Kullback-Leibler divergence term that assesses the closeness of the normally distributed latent space variables to the standard normal distribution. 
In contrast, the orthogonal autoencoder (OAE)~\cite{wang:2019}, originally introduced for clustering tasks, enforces orthogonality of the latent variables in the loss function in a deterministic manner. 
Again, there is a hyperparameter ($\lambda>0$) that controls the importance of orthogonality in comparison to the reconstruction quality in the training process. 
In~\cite{CACCIARELLI2022107853}, the OAE was used to provide uncorrelated latent variables for a statistical process control framework. 

Work on disentangling the latent variables in the context of low-dimensional representations of fluid dynamic fields has been focusing on $\beta$-VAEs. 
Eivazi et al.~\cite{eivazi:2022} proposed using a $\beta$-VAE for the extraction of near-orthogonal modes of turbulent flows. More recently, Solera-Rico et al.~\cite{solera-rico:2024} combined this approach with a transformer to model the dynamics of fluid flows in the obtained disentangled latent space. 
Kang et al.~\cite{kang:2022} employ a $\beta$-VAE in combination with a Gaussian process regression to construct a reduced-order model (ROM) for transonic flows. After dimension reduction, the input parameters of the full-order simulation are mapped to the disentangled latent space using  regression techniques similar to other works~\cite{Swischuk.2019,Agostini20,Pache22,lazzara:2022, DIASRIBEIRO2023105949} to obtain a more interpretable surrogate model. They demonstrate the ability of the $\beta$-VAE to use only a few physics-aware latent variables, when trained with a larger than needed latent space dimension. Notice that the phenomenon of VAEs having inactive latent variables was reported before, e.g.,~\cite{bowman2016generating, burda2016importanceweightedautoencoders, ladder_vae}.

In this work, we compare the established $\beta$-VAE with two non-probabilistic frameworks, the OAE and an approach we call uncorrelated autoencoder (UAE), which directly uses the correlation matrix of the latent variables to enforce the desired property of a disentangled latent space. 
Slightly similar proposals to complement the loss function have recently been used for an image style transfer application, where correlations between different channels of encoded feature maps are reduced~\cite{KIM2021148}, as well as for the uncorrelated sparse autoencoder, pursuing other encoding goals~\cite{Savargaonkar:2024}.
  
Two data sets are used in this work. The first example is a benchmark example for periodic flows that has already been used in publications of similar strategies for disentangled latent space ROM. The second case contains aircraft ditching load data and is used as an industrial application in this work. 
Ditching is the emergency landing on water, and its analysis is critical for the certification of a commercial aircraft. To reduce the computational effort, simulation-based ditching analysis is usually performed using a one-way coupling between the fluid and the structure, i.e., in a first step the hydrodynamic loads are calculated and subsequently the structural analysis is performed. 
In the long-term, the current research aims to include approximate deformations to the calculation of the hydrodynamic loads with ML. 
First steps towards the spatio-temporal prediction of ditching loads using combinations of convolutional autoencoder (CAE) with LSTM networks and Koopman operator approaches that can include deformation caused changes of the loads are documented in~\cite{schwarz:2024}.  
Interpretabilty of the data-driven surrogate model is obviously critical in an industrial process. The goal within this work is to advance the employed ML methods in this regard.
To the best of our knowledge, this is the first work for applications related to fluid dynamics that applies a loss function focusing on the correlation matrix of the latent variables and compares the OAE to the $\beta$-VAE and the UAE.

The remainder of the paper is structured as follows: In Sec.~\ref{sec:2}, the used autoencoder frameworks are briefly presented. Results on both datasets are shown in Sec.~\ref{sec:3} and conclusions are drawn in Sec.~\ref{sec:4}.

\section{Methods}
\label{sec:2}

\begin{figure}[!h]
     \centering
     \includegraphics{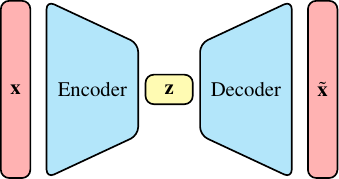}
    \caption{Basic autoencoder scheme.}
    \label{fig:ae}
\end{figure}

An autoencoder comprises an encoder and a decoder, as illustrated in Fig.~\ref{fig:ae}. The encoder maps the input $\mathbf{x}\in\mathbb{R}^n$ to a usually lower dimensional latent space representation $\mathbf{z}\in\mathbb{R}^m$ and the decoder subsequently maps  $\mathbf{z}$ to the full dimension to obtain a reconstruction of the input, $\tilde{\mathbf{x}}\in\mathbb{R}^n$. Encoder and decoder usually have symmetric structure. 
All models in this work are based on a CAE, as the considered physical data is two-dimensional in space and convolutional layers are usually better than fully connected layers at capturing spatial dependencies.

  A standard autoencoder reconstruction loss comparing input $\mathbf{x}$ and the reconstruction $\tilde{\mathbf{x}}$, i.e., mean squared error (MSE), is considered for all models so that encoder and decoder can be used to decrease or increase the spatial dimension of the data at hand. 
In this work, the goal of the reduced order models is to obtain disentangled latent variables $\mathbf{z}$ while keeping the reconstruction error as low as possible. To this end, the three models considered, i.e., $\beta$-VAE, OAE and UAE, each use a different second term in the loss function that acts in the latent space to effect the disentanglement. 

\subsection{Orthogonal Autoencoder (OAE)}
The orthogonal autoencoder (OAE) was introduced by Wang et al.~\cite{wang:2019} for clustering tasks. In addition to the reconstruction loss, the OAE enforces the orthogonality of the latent variables within the loss function for a mini batch of size $b$
\begin{align}
    \mathrm{Loss_{OAE}}(\mathbf{X},\tilde{\mathbf{X}}) =
    \frac{1}{b\cdot n}\lVert \mathbf{X} - \tilde{\mathbf{X}} \rVert^2_\mathrm{F} + 
    \frac{\lambda}{m^2}\lVert \mathbf{Z}^T \mathbf{Z} - \mathbf{I} \rVert^2_\mathrm{F}, \quad \lambda>0,
    \label{eq:oae_loss}
\end{align}
where $\mathbf{X}\in \mathbb{R}^{b\times n}$ contains the inputs for one mini batch, $\mathbf{Z} \in \mathbb{R}^{b\times m}$ is a matrix containing the latent vectors of a mini batch as rows and $\mathbf{I} \in \mathbb{R}^{m\times m}$.

\subsection{Uncorrelated Autoencoder (UAE)}
To measure the independence of the latent variables, the Pearson correlation matrix $\mathbf{R} \in \mathbb{R}^{m\times m}$ of the latent vectors, and its determinant $\det(\mathbf{R})$ are often considered~\cite{eivazi:2022, kang:2022, CACCIARELLI2022107853, solera-rico:2024}. The matrix is computed as follows
\begin{align*}
    \mathbf{R}_{ij} = \frac{\sum^K_{k=1}(z_i^k-\bar{z_{i}})(z_j^k-\bar{z_j})}{\sqrt{\sum^K_{k=1}(z_i^k-\bar{z_{i}})^2\sum^K_{k=1}(z_j^k-\bar{z_{j}})^2}}, \quad 1\le i,j \le m,
\end{align*}
where $K$ is the number of considered training or validation examples used for the calculation, $z_i^k$ denotes the value of the $i$th latent variable on the $k$th data example, and $\bar{z_{i}}$ is the mean value of the $i$th latent variable over all $k$ examples.
The matrix $\mathbf{R}$ is symmetric and has unit values on the diagonal. 
A correlation matrix close to the identity matrix and thus a determinant close to one corresponds to a good level of disentanglement. 

The second non-probabilistic approach considered here, which we term uncorrelated autoencoder (UAE), uses this criterion directly in the loss function. The employed loss function is similar to the correlation losses in~\cite{KIM2021148} and~\cite{Savargaonkar:2024}, where the absolute values of the non-diagonal entries of the correlation matrix are forced to be small. The present study follows the OAE method and uses the squared entries
\begin{align}
    \mathrm{Loss_{UAE}}(\mathbf{X},\tilde{\mathbf{X}}) =  \frac{1}{b\cdot n}\lVert \mathbf{X} - \tilde{\mathbf{X}} \rVert^2_\mathrm{F} + 
    \frac{\nu}{m^2}\lVert \mathbf{R} - \mathbf{I} \rVert^2_\mathrm{F}, \quad \nu>0,
    \label{eq:our_loss}
\end{align}
where $\mathbf{R} \in \mathbb{R}^{m\times m}$ is the Pearson correlation matrix of the latent vectors of a mini batch of size $b$ and $\mathbf{I} \in \mathbb{R}^{m\times m}$.
The UAE therefore focuses directly on the two properties of interest, reconstruction quality and disentanglement, without forcing the latent space to follow a normal distribution as the ($\beta$-)VAE presented next.  

\subsection{Variational Autoencoder (VAE)}
\begin{figure}[!h]
     \centering
    \includegraphics{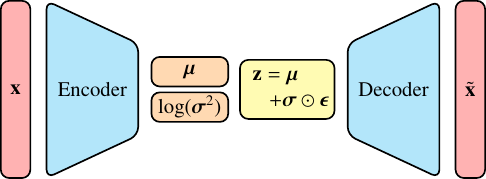}
    \caption{Scheme of a VAE.}
    \label{fig:vae}
\end{figure}

The latent space representations in a VAE are forced to follow a probability distribution, which is usually assumed to be normally distributed: $\mathbf{z}\sim \mathcal{N}(\boldsymbol{\mu},\boldsymbol{\Sigma})$, with mean $\boldsymbol{\mu}\in \mathbb{R}^m$ 
and assumed diagonal covariance matrix $\boldsymbol{\Sigma}\in\mathbb{R}^{m\times m}$ where the diagonal entries are $(\sigma_i)^2, i\in \{1,\dots,m\}$. 
 The encoder provides the two outputs $\boldsymbol{\mu}$ and 
$\log (\boldsymbol{\sigma}^2)$, as illustrated in Fig.~\ref{fig:vae}, 
where for stability reasons the $\log$-variance $\log (\boldsymbol{\sigma}^2)$ is returned. 
Using $\boldsymbol{\sigma} = \exp(0.5\cdot\log (\boldsymbol{\sigma}^2))$ and  generating $\boldsymbol{\epsilon} \sim \mathcal{N}(\mathbf{0},\mathbf{I})$, the latent variables $\mathbf{z} = \boldsymbol{\mu} + \boldsymbol{\sigma} \odot \boldsymbol{\epsilon}$ are computed, which are the input for the decoder. This is the \emph{reparameterization trick}~\cite{Kingma:2013} to enable classical backpropagation, as $\mathbf{z}$ can be considered a deterministic variable during training. The gradient with respect to the encoder can otherwise cause problems \cite{Kingma:2013}.
The loss function of a VAE is composed of a reconstruction loss, as well as a term that forces the posterior distribution in the latent space to be close to the chosen prior distribution. In this work, these are the MSE and the Kullback-Leibler (KL) divergence, which forces the posterior $\mathcal{N}(\boldsymbol{\mu},\boldsymbol{\Sigma})$ to be close to the prior $\mathcal{N}(\mathbf{0},\mathbf{I})$. The formula for the loss is
\begin{align}
    \mathrm{Loss_{VAE}}(\mathbf{X},\tilde{\mathbf{X}}) 
     =\frac{1}{b\cdot n}\lVert \mathbf{X} - \tilde{\mathbf{X}} \rVert^2_\mathrm{F} + \frac{\beta}{b\cdot 2}\sum_{i=1}^b\sum_{j=1}^m (\sigma_{ij}^2+\mu_{ij}^2 - 1 - \log(\sigma_{ij}^2)), \quad \beta>0,
    \label{eq:vae_loss}
\end{align}
where $b$ is the mini batch size, $\mathbf{X}\in \mathbb{R}^{b\times n}$ contains the inputs for one mini batch, and $\mu_{ij}$ and $\sigma^2_{ij}$ denote the $j$th component of mean and variance for the $i$th example in the mini batch. 
For $\beta= 1$, this is referred to as the VAE \cite{Kingma:2013} and $\beta\neq 1$ denotes the $\beta$-VAE~\cite{Higgins2016betaVAELB}.

The latent space obtained from VAEs or $\beta$-VAEs is usually continuous. That is a useful feature for the generation of new data, for which samples of the probability distribution in the latent space are decoded. 
Because $\boldsymbol{\epsilon}\sim \mathcal{N}(\mathbf{0},\mathbf{I})$ is generated every time the VAE is used to encode data, the trained model is not deterministic. 
Mind that in the work of Solera-Rico et al.~\cite{solera-rico:2024}, the trained $\beta$-VAE is applied in a deterministic manner, as only the mean values $\boldsymbol{\mu}$ are used by the models that learn the dynamics of fluid flows. 
Using a deterministic model from the beginning could, however, be the simpler approach as the training of VAEs can be challenging.

A common challenge is the so-called \emph{posterior collapse}, in which the posterior matches the prior distribution and the decoder ignores the latent vector $\mathbf{z}$~\cite{bowman2016generating, ladder_vae, lucas2019understanding}. In that case, the model does not learn meaningful representations of the inputs $\mathbf{x}$. Although techniques exist to mitigate this problem, such as including a warm-up phase for the reconstruction loss in the beginning of the training in which $\beta$ is initially set to 0 and then gradually increased~\cite{bowman2016generating, ladder_vae}, or using a different prior distribution, e.g., \cite{NIPS2017_7a98af17, davidson2022hypersphericalvariationalautoencoders}, this requires additional consideration for the training of VAEs. 

On the contrary, the collapse or inactiveness of several \emph{single} latent variables $z_i$, which do not help the reconstruction, is a beneficial feature for a disentangled latent space~\cite{burgess2018understandingdisentanglingbetavae} and for surrogate modeling, as this might help to identify the physics-aware latent variables~\cite{kang:2022}.

\subsection{Model Comparison}
The following section attempts to explain the similarities and differences between the three loss function of the three autoencoder models.
Let $\mathbf{Z}\in\mathbb{R}^{b\times m}$ contain the $b$ latent space vectors of a mini batch, and $\mathbf{z}_j$, $j=1,\ldots,m$ denote the columns of $\mathbf{Z}$. By initialization of the networks, for which typically a uniform distribution centered at zero is used, and the loss functions used, we can assume that all $m$ latent space features have approximately zero mean and a nonzero variance. If the features have zero mean, $\|\mathbf{z}_j\|^2=\langle\mathbf{z}_j,\mathbf{z}_j\rangle=b\textsf{Var}(z_j^k)$ for all $j=1,\ldots,m$.
The OAE enforces nearly orthonormal columns of $\mathbf{Z}$, which also affects the variance, as the columns are scaled to have approximately unit length.
The KL divergence in the ($\beta$-)VAE shifts the latent variables towards zero mean values with unit variances.
A combination of both, i.e., latent variables with a zero mean and (scaled) orthonormal columns, results in the minimization of the latent space loss function employed by the UAE:
with $i\neq j$, $\mathbf{z}_i^T\mathbf{z}_j=0$ and $\bar{z}_{i}=\bar{z}_{j}=0$. Then $\sum^b_{k=1}(z_i^k-\bar{z}_{i})(z_j^k-\bar{z}_j) = \sum^b_{k=1}z_i^kz_j^k=0 \Rightarrow \mathbf{R}_{i,j}=0$.
Vice versa, an identity correlation matrix $\mathbf{R}=\mathbf{I}$ is obtained exactly, if the columns of $\mathbf{Z}$ shifted by their mean are nonzero orthogonal. By assumption the mean will be approximately zero, thus it makes sense to directly minimize the UAE loss.

\subsection{Analysis of Latent Variables}
Having disentangled latent variables, the impact of the different variables may be analyzed. This can either be done by decoding latent vectors and analyzing the reconstruction, or solely in the latent space by encoding data from the full dimensional space. 

Eivazi et al.~\cite{eivazi:2022} use the former approach in conjunction with a $\beta$-VAE. They map full order data to the latent space and set all components except the component $z_i$ to zero. Subsequently, the resulting vectors are decoded to obtain the $i$th modes, and the whole procedure is repeated for all latent variables. Finally, the obtained energy from those reconstructions is calculated and the mode providing the largest energy is identified. If the largest energy is obtained using the $j$th latent variable, this $j$th mode is ranked as the first mode. This procedure is then repeated to determine the second mode in the ranking, but without setting $z_j$ to zero, and so on.
 Kang et al.~\cite{kang:2022} perform their analysis of the $\beta$-VAE latent variables completely in the latent space, which avoids the application of the decoder. They calculate the KL divergence of each latent variable to observe which variables contain useful information. Additionally, they consider the standard deviation of each latent variable to assess, which variables are active [inactive] and thus change [do not change] for varying inputs.  

\subsubsection{Mode Generation}
Single latent variables can be associated with modes. To this end, the latent variables are ranked as described above to identify the most active latent variables. {Inspired by Kang et al.~\cite{kang:2022}, this ranking is performed inside the latent space using the standard deviation and KL divergence of the latent variables.} 
The spatial modes are generated by decoding the $i$th unit vector of the latent space dimension for some of the highest ranked $i$ ~\cite{eivazi:2020, solera-rico:2024}. 
However, there is a non-linear behavior in the output when changing the value of $z_i$, cf. supplementary information of~\cite{solera-rico:2024}. 
 For this reason, when analyzing the impact of a latent variable, we do not look at a single value for that latent variable, but rather at a range of values while keeping the other latent variables fixed.
\section{Results}
\label{sec:3}
We first verify the performance of the three autoencoder models
 in enforcing a disentangled latent space for model reduction of the data extracted from a periodic flow benchmark case suggested by Solera-Rico et al.~\cite{solera-rico:2024, solera_rico_2024_dataset}. 
Subsequently, we apply all models to reduce the load data extracted from aircraft ditching simulations. This more complex industrial test case uses the same data recently used in~\cite{schwarz:2024}. 
All models are implemented in PyTorch~\cite{NEURIPS2019_9015} and trained on a workstation equipped with an NVIDIA GeForce RTX 3090 GPU.

\subsection{2D Periodic Benchmark Flow}
\label{sec:periodicflow}
\begin{figure}[!h]
     \centering
    \includegraphics{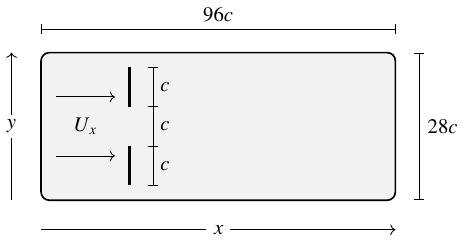}
    \caption{{Illustration of the 2D periodic flow configuration employed for the verification analysis \cite{solera-rico:2024, Asztalos:2024, solera_rico_2024_dataset}}.}
    \label{fig:periodic_flow_scheme}
\end{figure}

A 2D benchmark case suggested in~\cite{solera-rico:2024} {and previously described in~\cite{Asztalos:2024}} is used for the verification of the models. The configuration is illustrated in Fig. \ref{fig:periodic_flow_scheme}.
The simulation data includes 1000 equidistant
time steps of a periodic flow over two flat plates of length $c$ 
whose centers are offset by $2c$ in the vertical ($y$-) direction
 and approached at an angle of attack of 90°. The first 900 time steps are used for training and the subsequent 100 time steps are used for validation.
The homogeneous, unidirectional horizontal inflow velocity $U_x$ results in a Reynolds number of $Re_{c,U_x}=40$. 
 The computational domain is of size $96c\times28c$, and the spatial resolution of the data refers to $300 \times 88$ homogeneously spaced, equidistant  
grid points in the horizontal ($x$-) and vertical ($y$-) direction, respectively. The temporal resolution of the data refers to $\Delta t= c/U_x$. 

Attention will be restricted to the reduction of the horizontal and vertical velocities by the autoencoders.  The data of this case is particularly suitable for comparison between machine learning and classical model reduction strategies, since it can be accurately represented with a latent space dimension of $m=2$. Therefore, the flow field was previously used to compare the results obtained from a $\beta$-VAE with a POD approach~\cite{solera-rico:2024}. The ability of the models to reduce to two modes and the behavior of the two modes is the focus of this subsection. To safe space, we limit ourselves to the reconstruction of the 11th snapshot of the validation cycle as an exemplary reference snapshot.

\subsubsection{Autoencoder Configuration}
\label{sec:periodicaeconf}
The employed $\beta$-VAE agrees with that of Solera-Rico et al.~\cite{solera-rico:2024}, and the decoder structure of the $\beta$-VAE was also used for the decoder of the OAE and the UAE, see Table~\ref{tab:cae_periodic_flow}. 
The encoder structure of the OAE and UAE is generally consistent with that of the $\beta$-VAE, with one exception being the last layer, which has been modified to have $m$ neurons instead of $2\cdot m$, since unlike the $\beta$-VAE, the OAE and UAE do not require two outputs for the mean and log-variance.
Similar to Solera-Rico et al.~\cite{solera-rico:2024}, \textsf{ELU}~\cite{clevert2016fastaccuratedeepnetwork} is the activation function used for the hidden layers in the encoder and the decoder. The output layers of the encoder and the decoder are linear.

\begin{table}[h!]
\caption{Structure of encoder and decoder. The kernel size is $3\times 3$, and the stride is 2 for all Conv2d and ConvTranspose2d layers. For the $\beta$-VAE, one Linear(4) layer representing $\boldsymbol{\mu}$ and $\log (\boldsymbol{\sigma}^2)$ and the calculation of $\mathbf{z} = \boldsymbol{\mu} + \boldsymbol{\sigma} \odot \boldsymbol{\epsilon}$ replace the single Linear(2) layer in the encoder.}
\centering
\begin{tabular}{cccc}
    \hline
    Encoder & Output shape & Decoder & Output shape\\
    \hline
 Input & (300,88,2) & Linear(256) & (256)\\ 
     Conv2d(out\_channels=8) & (150,44,8) & Linear(2560) & (2560)\\
     Conv2d(out\_channels=16) & (76,22,16) & Unflatten(5,2,256) & (5,2,256)\\
     Conv2d(out\_channels=32) & (38,12,32) & ConvTranspose2d(out\_channels=128) & (10,4,128) \\
     Conv2d(out\_channels=64) & (20,6,64) &  ConvTranspose2d(out\_channels=64) & (20,6,64) \\
     Conv2d(out\_channels=128) & (10,4,128) & ConvTranspose2d(out\_channels=32) & (38,12,32) \\
     Conv2d(out\_channels=256) & (5,2,256) & ConvTranspose2d(out\_channels=16) & (76,22,16)\\
     Flatten() & (2560)  & ConvTranspose2d(out\_channels=8) & (150,44,8)\\  
     Linear(256) & (256) & ConvTranspose2d(out\_channels=2) & (300,88,2) \\
     Linear(2) & (2) \\
    \hline
\end{tabular}
\label{tab:cae_periodic_flow}
\end{table}

The models are trained using Adam~\cite{Kingma:2015} with a mini batch size of 256. The training employs 1000 epochs using a 1-cycle learning rate~\cite{smith:2018} that starts at $1\cdot 10^{-4}$, increases to $2\cdot 10^{-4}$ until the 200th epoch and subsequently decreases to $5\cdot 10^{-6}$ until the 1000th epoch as in~\cite{solera-rico:2024}. In all cases, no pretraining for the autoencoder with just the reconstruction loss is performed, meaning reconstruction loss and latent space loss are both used in every epoch.

\subsubsection{Influence of Loss Function Weights}
\label{sec:periodicflow_loss_function_weights}
To compare how the models perform when changing the weight for the disentanglement contribution to the loss function, we train each model with 6 different weighting values $10^{-5}$, $10^{-4}$, $10^{-3}$, $10^{-2}$, $10^{-1}$ and $10^{0}$ for $\beta$ (VAE), $\lambda$ (OAE) and $\nu$ (UAE), respectively. 
Note that for the $\beta$-VAE, the value $\beta=10^{-3}$ used herein does not correspond to the value of $\beta=10^{-3}$ also used in~\cite{solera-rico:2024}, since the KL divergence was divided by the latent space dimension of $m=2$. The latter is concluded from the accompanying code of~\cite{solera-rico:2024}. Our value of $\beta=10^{-3}$ is thus two times larger. 

\begin{figure}[h!]
    \centering
    \includegraphics[width=0.8\textwidth]{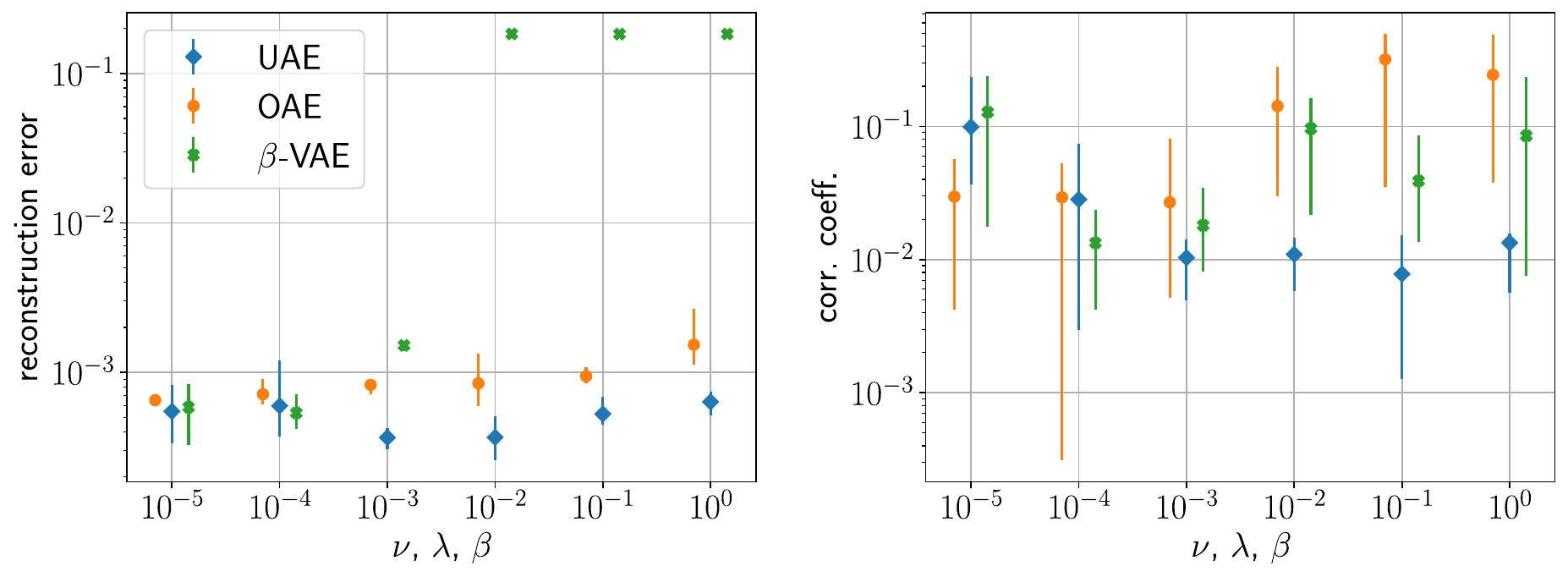}
    \caption{Reconstruction errors obtained for the validation set and resulting absolute values of correlation coefficients between the two latent variables returned by the three different autoencoder models. The errorbars correspond to average, lowest and highest values obtained from five trainings.}
    \label{fig:1cycleLR_loss_factors_vs_rec_losses_corr_coeff}
\end{figure}

We compare the results in terms of (a) the achieved reconstruction losses and (b) the achieved correlation coefficients between the two latent variables.
For this comparison, each model is trained five times for each of the six weighting values to include the influence of the randomness in the training process. 
Figure~\ref{fig:1cycleLR_loss_factors_vs_rec_losses_corr_coeff} shows the reconstruction errors (left) that are obtained on the validation set as well as the absolute values of the correlation coefficients between the two latent variables (right). 

In general, the variance of the reconstruction error between the five training runs is small for all autoencoders considered.
As the green symbols in the left graph show, the reconstruction error of the $\beta$-VAE was not minimized at all during training 
for the three larger $\beta$-values, i.e.,  $\beta=10^{0}, 10^{-1}, 10^{-2}$.  This indicates too high $\beta$-values that lead to a dominating behavior of the KL divergence. Reducing the weight to $\beta=10^{-3}$, this is not the case anymore and the average reconstruction error decreases by about two orders of magnitude. For $\beta=10^{-4}$, the error decreases even further and is similarly low as for $\beta=10^{-5}$.  {We also performed the same experiments with a warm-up phase for 50 epochs, in which $\beta$ is initially set to 0 and then gradually increased to the specific value. The warm-up strategy did return similar results, although the reconstruction loss  for $\beta=10^{-2}$  did minimize to values around $10^{-2}$ in a few cases. In combination with  $\beta=10^{-1}$ and $\beta=10^0$ the reconstruction errors obtained with a warm-up strategy at the end of the training  are at the same level as depicted in Fig. \ref{fig:1cycleLR_loss_factors_vs_rec_losses_corr_coeff}.}
The reconstruction errors of the other two models show a more robust behavior for the choice of weighting values, with the UAE (blue) delivering lower average reconstruction errors than the OAE (orange) in all cases presented. In none of the cases investigated did the reconstruction loss of the UAE and the OAE models remain at the initial level.

With regard to the correlation coefficients, we can observe a larger variance between the different training runs, especially for the OAE results displayed in orange. 
The UAE results displayed in blue yield the lowest average correlation coefficient for the four largest values of the weight, i.e., $\nu=10^{0}, 10^{-1}, 10^{-2}, 10^{-3}$. 
Although in these cases the second lowest value is returned by the $\beta$-VAE, this must be considered under the limitation that the reconstruction error was not minimized for the three largest values of $\beta$. 
For the two smallest weighting values, i.e., $10^{-4}$ and $10^{-5}$, the $\beta$-VAE and the OAE, respectively, yield the lowest average correlation coefficient, with the UAE providing the second lowest value for both weights.

\subsubsection{Mode Analysis for Small Latent Space Dimension}
\label{sec:modean-2Dm2}
To analyze the impact of the latent variables, we generate an exemplary snapshot, verify its reconstruction quality and subsequently change one latent variable, while keeping the other latent variable fixed. This procedure is done for both latent variables ($z_1, z_2$) and all three models, i.e., UAE, OAE and $\beta$-VAE.
Fixing the second latent variable can be done in two ways.
On the one hand, one can use the value corresponding to the exemplary snapshot, as in~\cite{Higgins2016betaVAELB, burgess2018understandingdisentanglingbetavae, kang:2022}. On the other hand, the other latent variable can simply be set to zero.
However, because the investigated periodic benchmark flow is quite simple and does not contain many significantly different states, we choose to keep the exemplary value of one latent variable and observe how the other latent variable changes the state. 

\begin{figure}[h!]
    \centering
    \includegraphics[width=\linewidth]{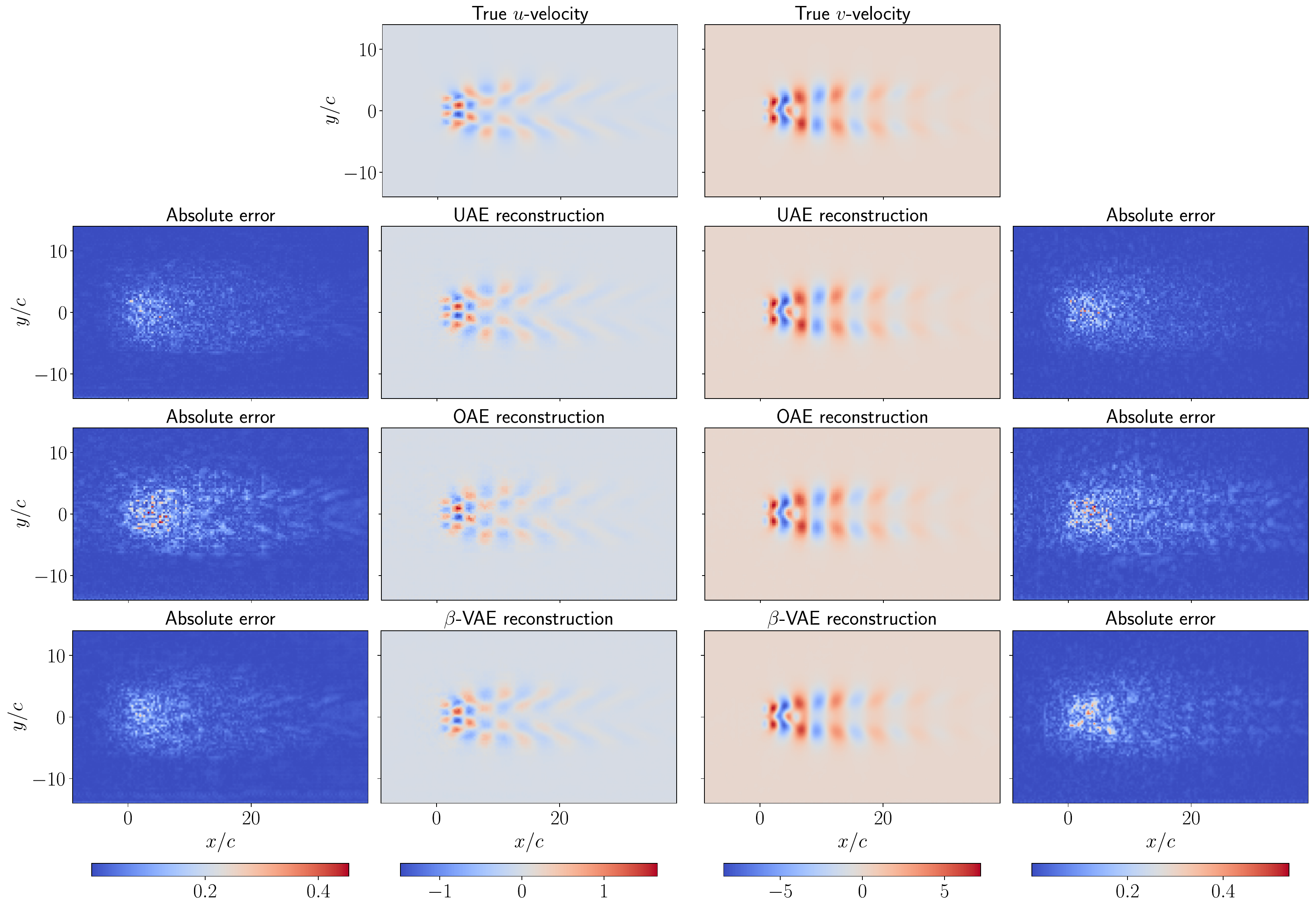}
    \caption{True $u$- and $v$-velocity fields, model reconstructions and related errors for the exemplary reference snapshot of the verification case.}
    \label{fig:case1_reconstruction}
\end{figure}

\begin{figure}[h!]
    \centering
    \includegraphics[width=0.8\textwidth]{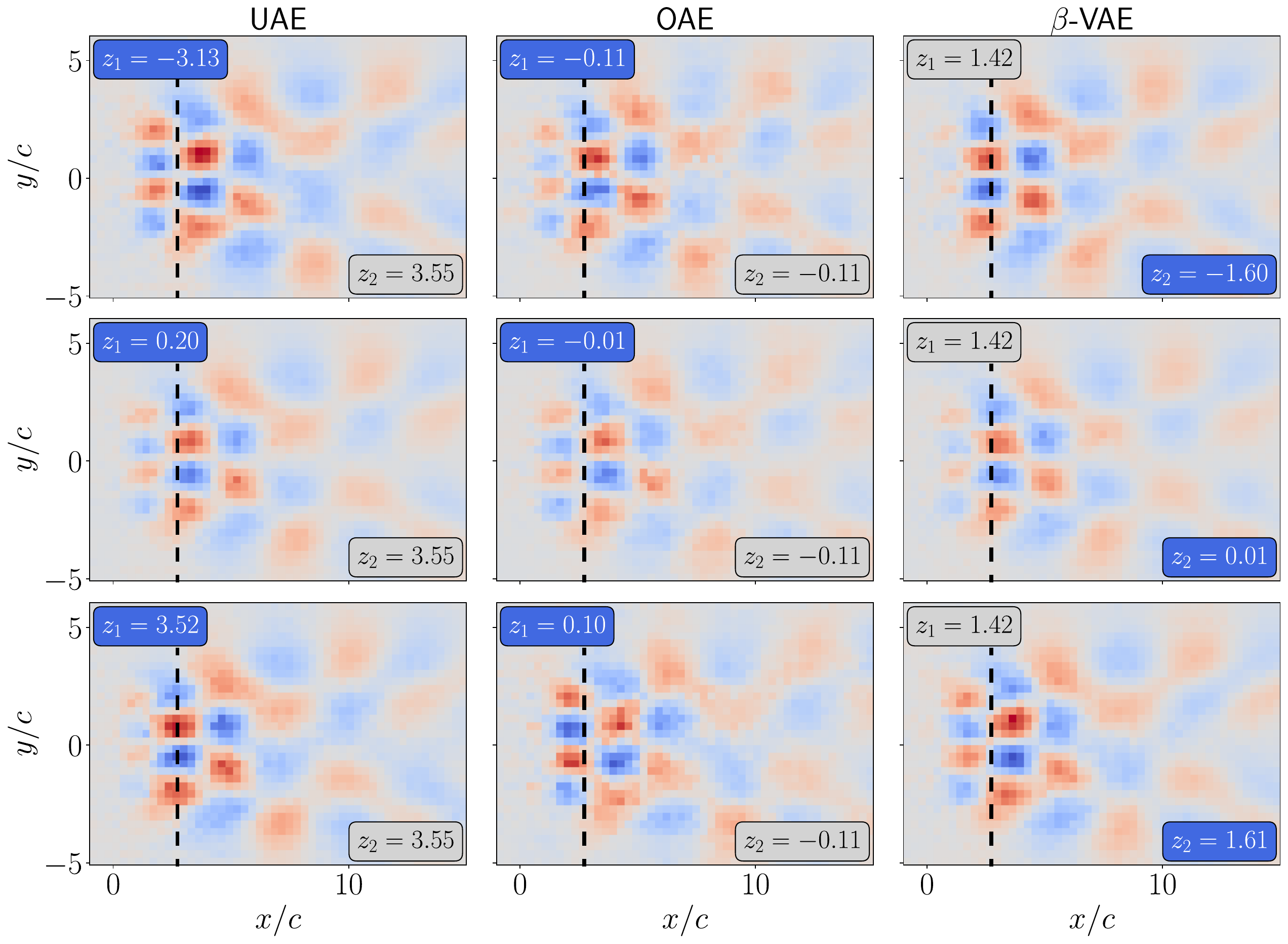}
    \caption{Impact of  changing $z_1$ for the UAE/OAE and $z_2$ for the $\beta$-VAE on the  $u$-velocity in the reference snapshot of the verification case using $m=2$ latent variables. The changed latent variable takes equidistant values ranging from the minimum to the maximum obtained for that variable on the validation set. The other latent variable is fixed to the corresponding latent space representation of the reference snapshot. The color code is the same for all examples, ranging from -1.87 (dark blue) to 1.90 (dark red).}
    \label{fig:case1_u_modes1}
\end{figure}

\begin{figure}[h!]
    \centering
    \includegraphics[width=0.8\textwidth]{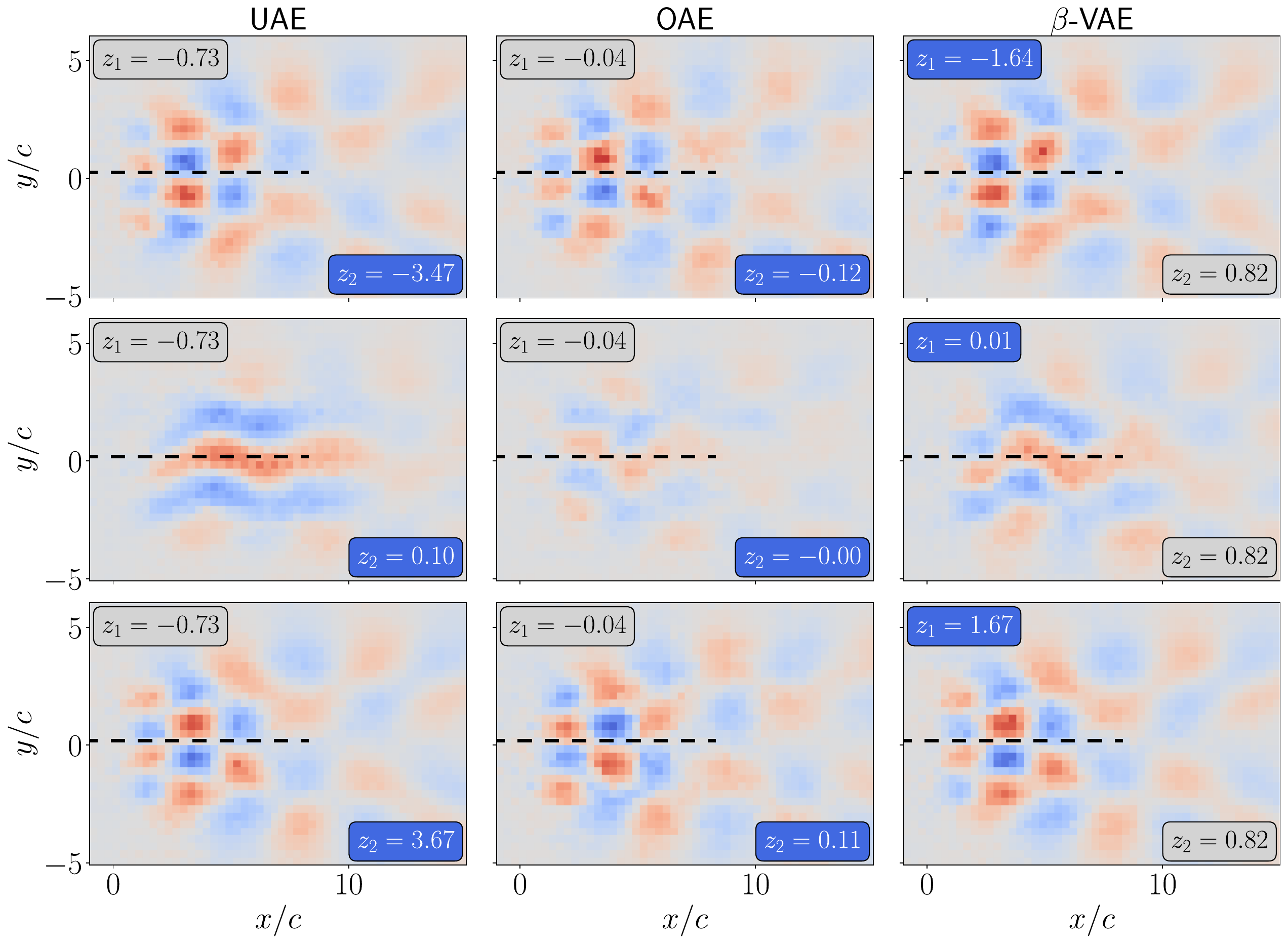}
    \caption{Impact of changing $z_2$ for the UAE/OAE and $z_1$ for the $\beta$-VAE on the $u$-velocity in the reference snapshot of the verification case using $m=2$ latent variables. The changed latent variable takes equidistant values ranging from the minimum to the maximum obtained for that variable on the validation set. The other latent variable is fixed to the corresponding latent space representation of the reference snapshot. The color code is the same for all examples, ranging from -1.87 (dark blue) to 1.90 (dark red).}
    \label{fig:case1_u_modes2}
\end{figure}

\begin{figure}[h!]
    \centering
    \includegraphics[width=0.8\textwidth]{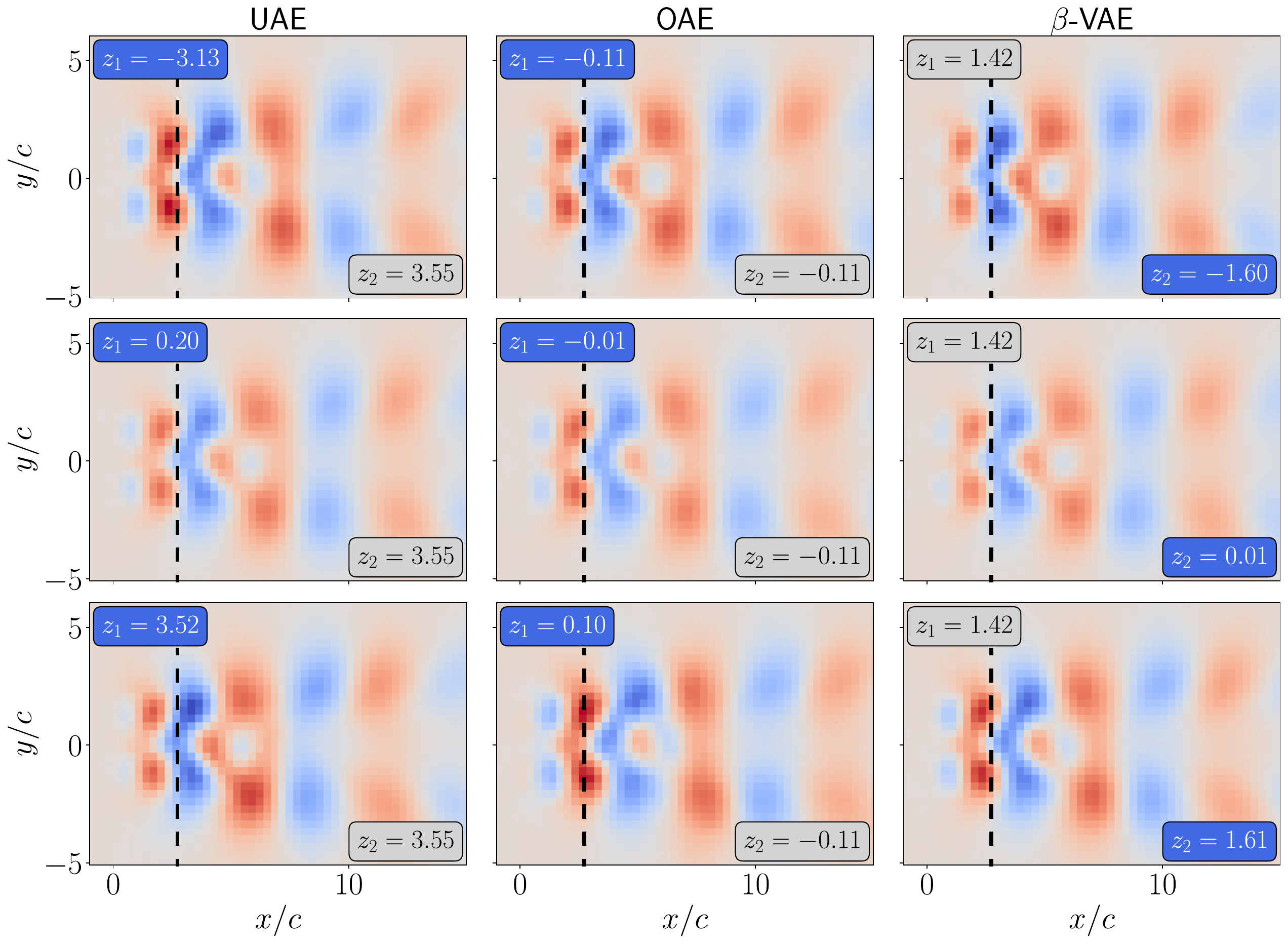}
    \caption{Impact of changing $z_1$ for the UAE/OAE and $z_2$ for the $\beta$-VAE on the $v$-velocity in the reference snapshot of the verification case using $m=2$ latent variables. The changed latent variable takes equidistant values ranging from the minimum to the maximum obtained for that variable on the validation set. The other latent variable is fixed to the corresponding latent space representation of the reference snapshot. The color code is the same for all examples, ranging from -11.58 (dark blue) to 10.41 (dark red).}
    \label{fig:case1_v_modes1}
\end{figure}

\begin{figure}[h!]
    \centering
    \includegraphics[width=0.8\textwidth]{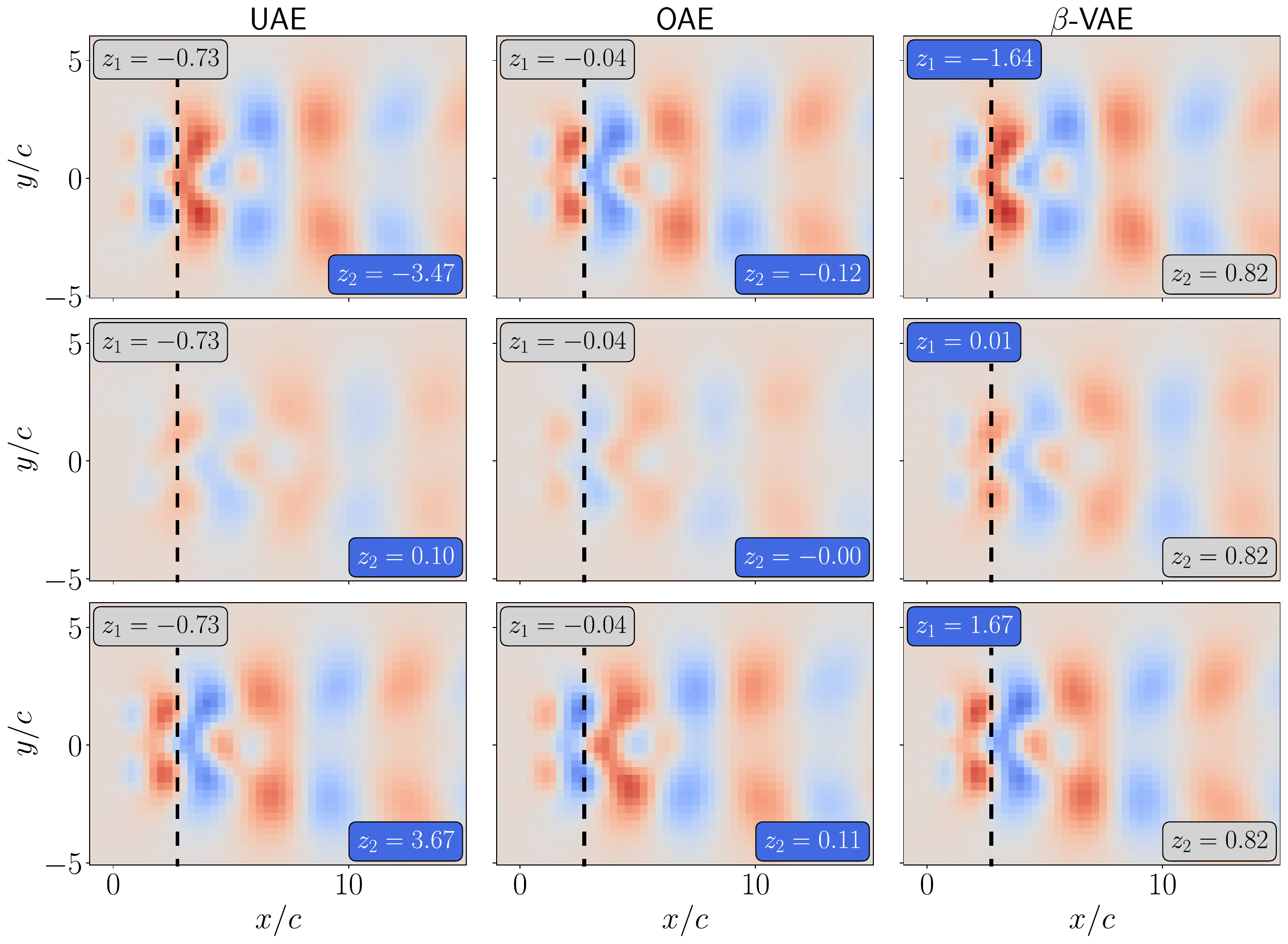}
    \caption{Impact of changing $z_2$ for the UAE/OAE and $z_1$ for the $\beta$-VAE on the $v$-velocity in the reference snapshot of the verification case using $m=2$ latent variables. The changed latent variable takes equidistant values ranging from the minimum to the maximum obtained for that variable on the validation set. The other latent variable is fixed to the corresponding latent space representation of the reference snapshot. The color code is the same for all examples, ranging from -11.58 (dark blue) to 10.41 (dark red).}
    \label{fig:case1_v_modes2}
\end{figure}

Figures~\ref{fig:case1_reconstruction}-\ref{fig:case1_v_modes2} display the baseline results obtained in combination with $\nu=10^{-2}$ for the UAE, and $\lambda=\beta=10^{-4}$ for the OAE and $\beta$-VAE for the  reference snapshot.
This choice of weights should ensure a satisfactory reconstruction error and a significant level of disentanglement.
 Figure \ref{fig:case1_reconstruction} compares the true values with the 
autoencoder-based reconstructions for the horizontal $u$-velocity (left) 
 and the vertical $v$-velocity (right) in an upstream portion of the domain. 
 The color coding used is indicated in the figure and is consistent for all $u$- and $v$-velocity images, respectively.
The figure reveals that only small reconstruction errors occur in conjunction with any of the three models for the chosen exemplary snapshot. 

 In Figs. \ref{fig:case1_u_modes1}-\ref{fig:case1_v_modes2}, one of the two latent variables, outlined by the dark background color, is changed, while the other latent variable, indicated by the light background color, remains unchanged.   The imposed changes of the variable values refer to the upper, lower and mean value of the entire value range.
Displayed results are confined to a spatial window ranging from {$-1\le x \le 15$ and $-5\le y\le 6$} to focus on the region of the most significant influences, and
the employed color codes agree.
Looking at the $u$-velocity depicted in Figs.~\ref{fig:case1_u_modes1}-\ref{fig:case1_u_modes2}, all models show a similar behavior. One latent variable, i.e., the first for the UAE and the OAE and the second for the $\beta$-VAE seems to induce a shift of the $u$-velocity pattern in horizontal ($x$-) direction, while the other seems to induce a shift in vertical ($y$-) direction including a corresponding wave shaped transition that is observed during the initial phase, cf. center graphs in Fig. \ref{fig:case1_u_modes2}. 
With attention given to the $v$-velocity depicted in Figs. \ref{fig:case1_v_modes1}-\ref{fig:case1_v_modes2}, no vertical displacement is observed in response to any changes of the latent variables. Instead, only horizontal shifts can be observed.
The changes induced by  $z_1$ (UAE, OAE) and $z_2$ ($\beta$-VAE) are much lower than in the other case.

\subsubsection{Physics Awareness for Higher Latent Space Dimension}
\label{sec:case1:higher_latent_dimension}
For the periodic flow at hand, two modes or latent variables are enough to capture (most of) the underlying physics~\cite{solera-rico:2024}. Solera-Rico et al.~\cite{solera-rico:2024} emphasize that a $\beta$-VAE with a larger latent space dimension $m>2$ still only finds two meaningful modes. This is in line with the findings by Eivazi et al.~\cite{eivazi:2022} and Kang et al.~\cite{kang:2022}, where the $\beta$-VAE is shown to find physics-aware latent variables. 
To assess the physics-awareness properties of the deterministic autoencoder models, we train all three models with a latent space dimension of $m=10$ and analyze whether they can learn using only two disentangled latent variables.  
Inspired by Kang et al.~\cite{kang:2022}, we employ the standard deviation 
(for all models) and the KL divergence (only for the $\beta$-VAE) of the latent variables to judge which latent variable  is active, i.e., physically significant. 
Subsequently, we generate and compare the modes obtained from the latent variables featuring the four highest standard deviations for an appropriate weighting value along the route outlined in Sec. \ref{sec:modean-2Dm2}.

\begin{figure}[h!]
    \centering
    \includegraphics[width=\textwidth]{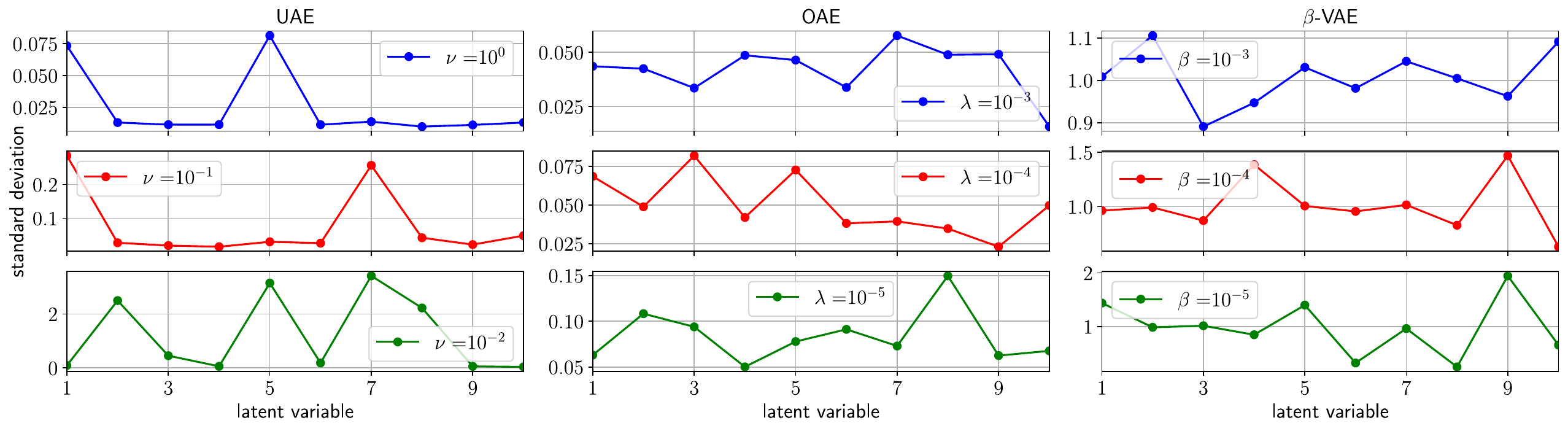}
    \caption{Standard deviations of latent variables for all three models obtained in the verification case for a latent space dimension of $m=10$. 
    Each model is trained with
    three different weights of the disentanglement contribution to the loss function.}
    \label{fig:case1_std}
\end{figure}

\smallskip
Figure~\ref{fig:case1_std} depicts the standard deviations of the considered ten latent variables in combination with three different weights of the disentanglement contribution to the loss function. 
The investigated weighting values were taken from the successful studies for $m=2$ latent variables, cf. Fig.~\ref{fig:1cycleLR_loss_factors_vs_rec_losses_corr_coeff}.  
 They refer to $\nu = 10^{0}$, $\nu = 10^{-1}$ and $\nu = 10^{-2}$ for the UAE, $\lambda = 10^{-3}$, $\lambda = 10^{-4}$ and $\lambda = 10^{-5}$ for the OAE, as well as $\beta = 10^{-3}$, $\beta = 10^{-4}$ and $\beta = 10^{-5}$ for the $\beta$-VAE.

\begin{figure}[h!]
    \centering
    \includegraphics[width=\textwidth]{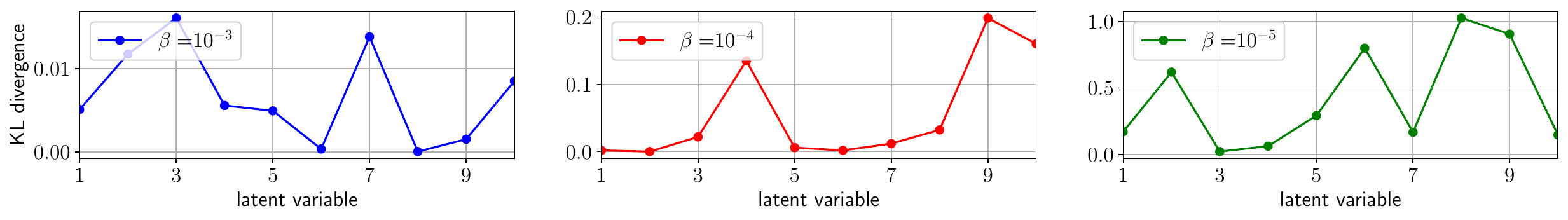}
    \caption{KL divergence of the latent variables for the $\beta$-VAE
    obtained in the verification case for a latent space dimension of $m=10$. 
    The model is trained with
    three different values of $\beta$ in the loss function.}
    \label{fig:case1_kld}
\end{figure}

Considering the UAE results on the left of Fig.~\ref{fig:case1_std}, we observe that for $\nu=10^0=1$ and $\nu=10^{-1}$ two latent variables --i.e., $z_1$ and $z_5$, as well as $z_1$ and $z_7$, respectively-- have  noticeably higher standard deviations than the others. In conclusion, these two latent variables change their values more for different inputs to the encoder and are therefore more active than the other latent variables. This in turn suggests that the model has successfully extracted the two essential variables required to describe this flow, although this remains to be verified by analyzing their modes.
For $\nu=10^{-2}$ the correlation penalty is less strongly involved during training, and four latent variables are more active than the rest. 
This behavior is probably explained by the {balance between the numerator and the denominator} inside the loss function~\eqref{eq:our_loss} forcing a latent variable to be close to its mean over all inputs. A smaller $\nu/m^2$ allows more active latent variables as they are less strongly forced to be close to their mean over all inputs. For ($m=10$, $\nu=1$) and 
($m=10$, $\nu=10^{-1}$), the MSE and the correlation penalty are balanced in a way that the UAE uses a sufficient amount of two latent variables for the reconstruction. 
For the OAE depicted in the center graphs of Fig.~\ref{fig:case1_std}, the standard deviations of the different latent variables are generally closer to each other. For the three displayed values of $\lambda$, two latent variables with a clearly higher standard deviation than the others cannot be identified, indicating that the OAE has difficulties to extract the two most crucial latent variables when trained with a latent space dimension of $m=10$ in combination with these $\lambda$ values. In combination with $\lambda=10^{-4}$ the largest standard deviation refers to $z_1, z_3$ and $z_5$.  %
Results for the $\beta$-VAE are shown in the right graphs of Fig.~\ref{fig:case1_std}. For the large value $\beta=10^{-3}$, the standard deviation is close to 1 and the KL divergence shown in Fig. \ref{fig:case1_kld} is approximately 0 for every latent variable. Moreover, 
 the latent variables with largest standard deviation and the largest KL divergence disagree and partly contradict each other.
This makes  $\beta=10^{-3}$ seem questionable in this case. 
For $\beta=10^{-4}$, $z_4$ and $z_9$ have a slightly higher standard deviation than the other latent variables, which mostly have standard deviations close to 1. Accordingly, the KL divergence for these two variables also increases.
Moreover, $z_{10}$ noticeably deviates from a unit standard deviation in conjunction with $\beta=10^{-4}$ and thus also displays an increased KL divergence. 
For $\beta=10^{-5}$, there are three latent variables {($z_1$, $z_5$ and $z_9$)} that are most active. Two latent variables $z_6$ and $z_8$ [$z_5$/$z_1$ and $z_9$] 
have a noticeably lower [higher] standard deviation than the rest. However, the KL divergence is visibly greater than 0 for many more variables, suggesting that $\beta$ is too small to separate the two crucial modes.

\begin{figure}[h!]
    \centering
    \includegraphics[width=0.8\textwidth]{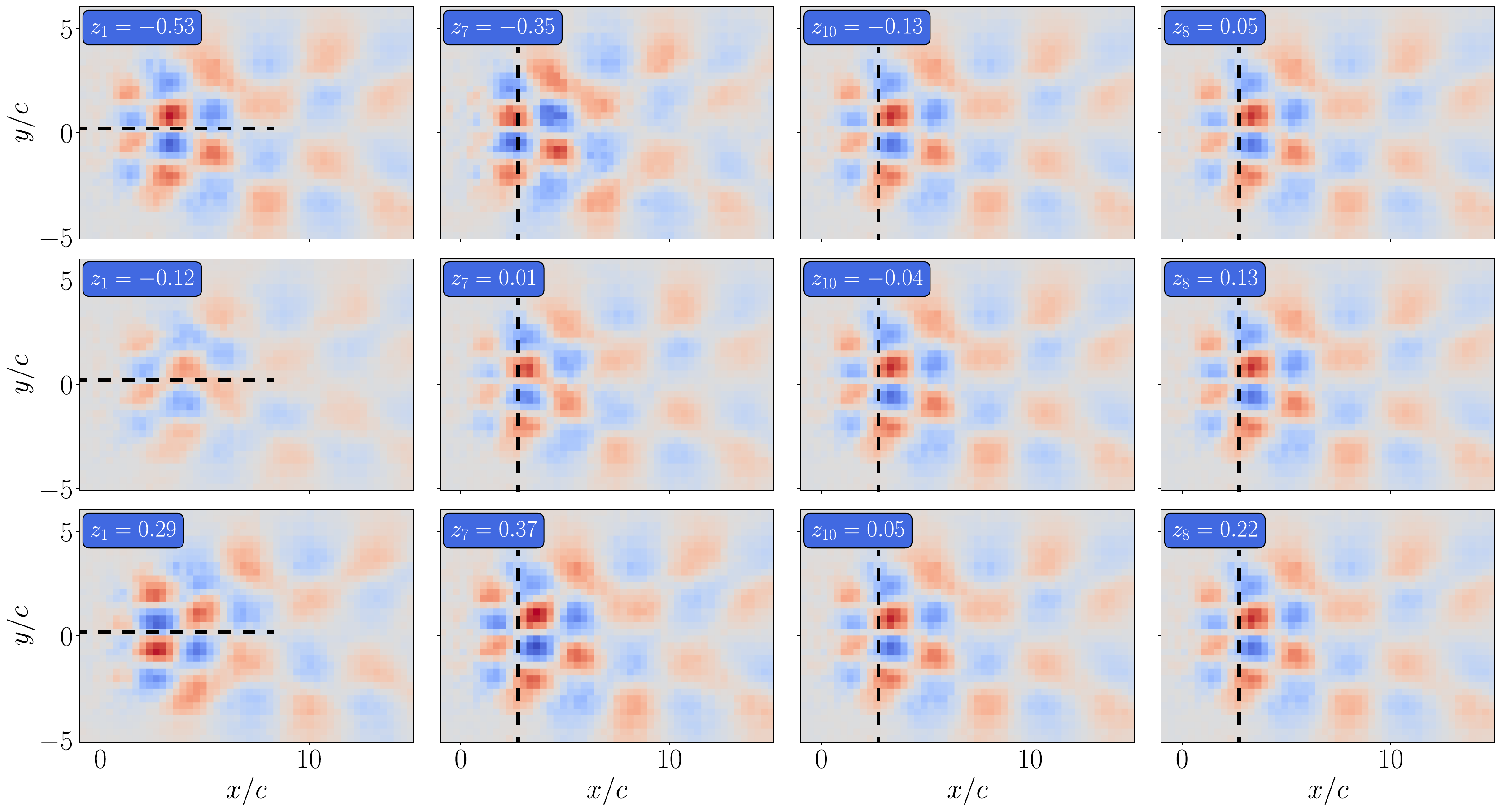}
    \caption{Impact of single latent variables on the $u$-velocity in the reference snapshot of the verification case for the UAE using a latent space dimension of $m=10$ and $\nu=10^{-1}$.  
    In each column, nine latent variables are fixed to the corresponding latent space representation of the snapshot and one variable takes equidistant values ranging from the minimum to the maximum value obtained
    for the validation set.
    Color code is the same for all examples in this figure, ranging from -1.65 (dark blue) to 1.66 (dark red).}
    \label{fig:Case1_UAE_Latent10_modes}
\end{figure}

\begin{figure}[h!]
    \centering
    \includegraphics[width=0.8\textwidth]{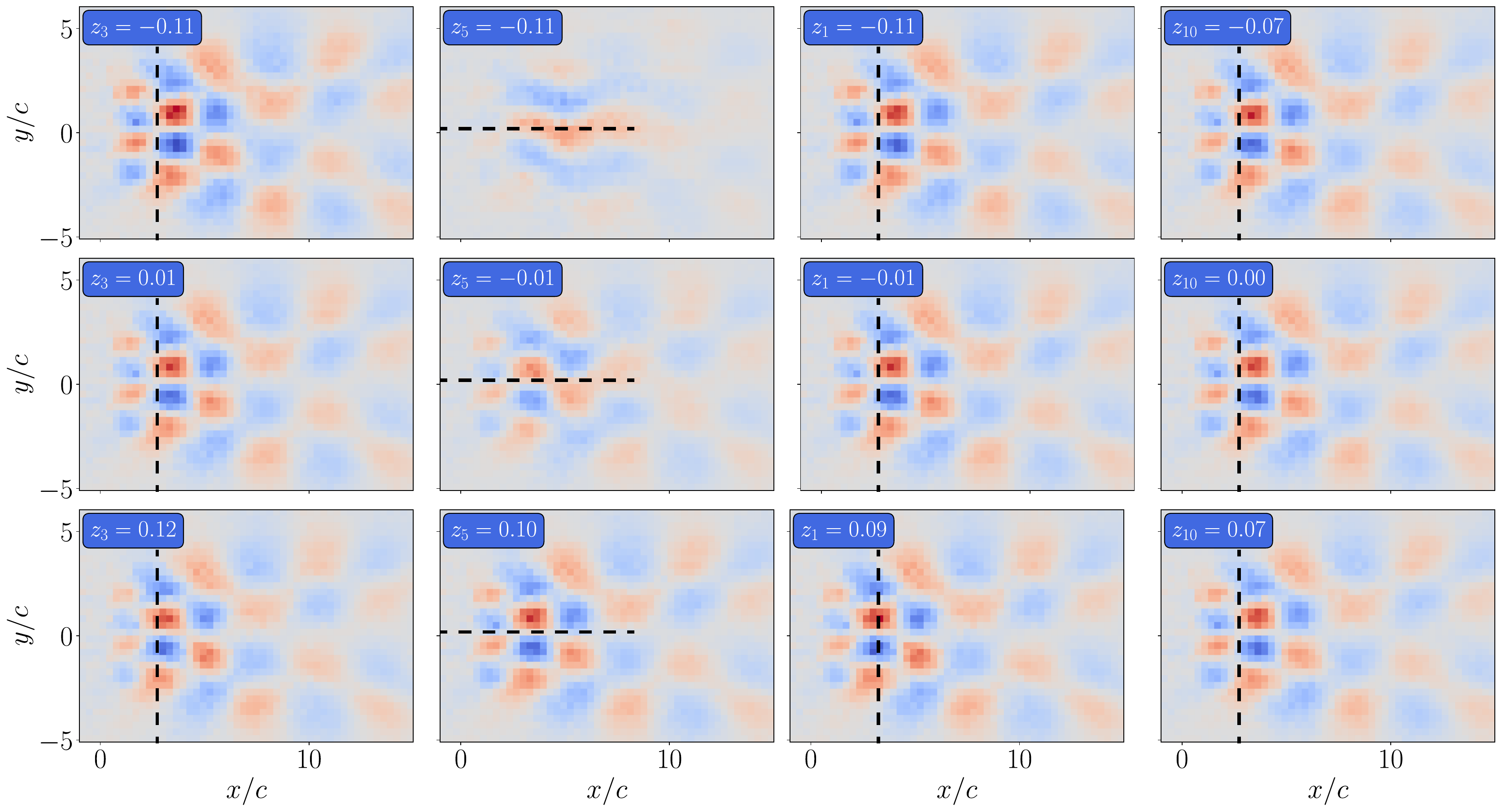}
    \caption{Impact of single latent variables on the $u$-velocity in the reference snapshot  of the verification case for the OAE using a latent space dimension of $m=10$ and $\lambda=10^{-4}$. 
    In each column, nine latent variables are fixed to the corresponding latent space representation of the snapshot and one variable takes equidistant values ranging from the minimum to the maximum value obtained
    for the validation set.
    Color code is the same for all examples in this figure, ranging from -1.65 (dark blue) to 1.74 (dark red).}
    \label{fig:Case1_OAE_Latent10_modes}
\end{figure}

\begin{figure}[h!]
    \centering
    \includegraphics[width=0.8\textwidth]{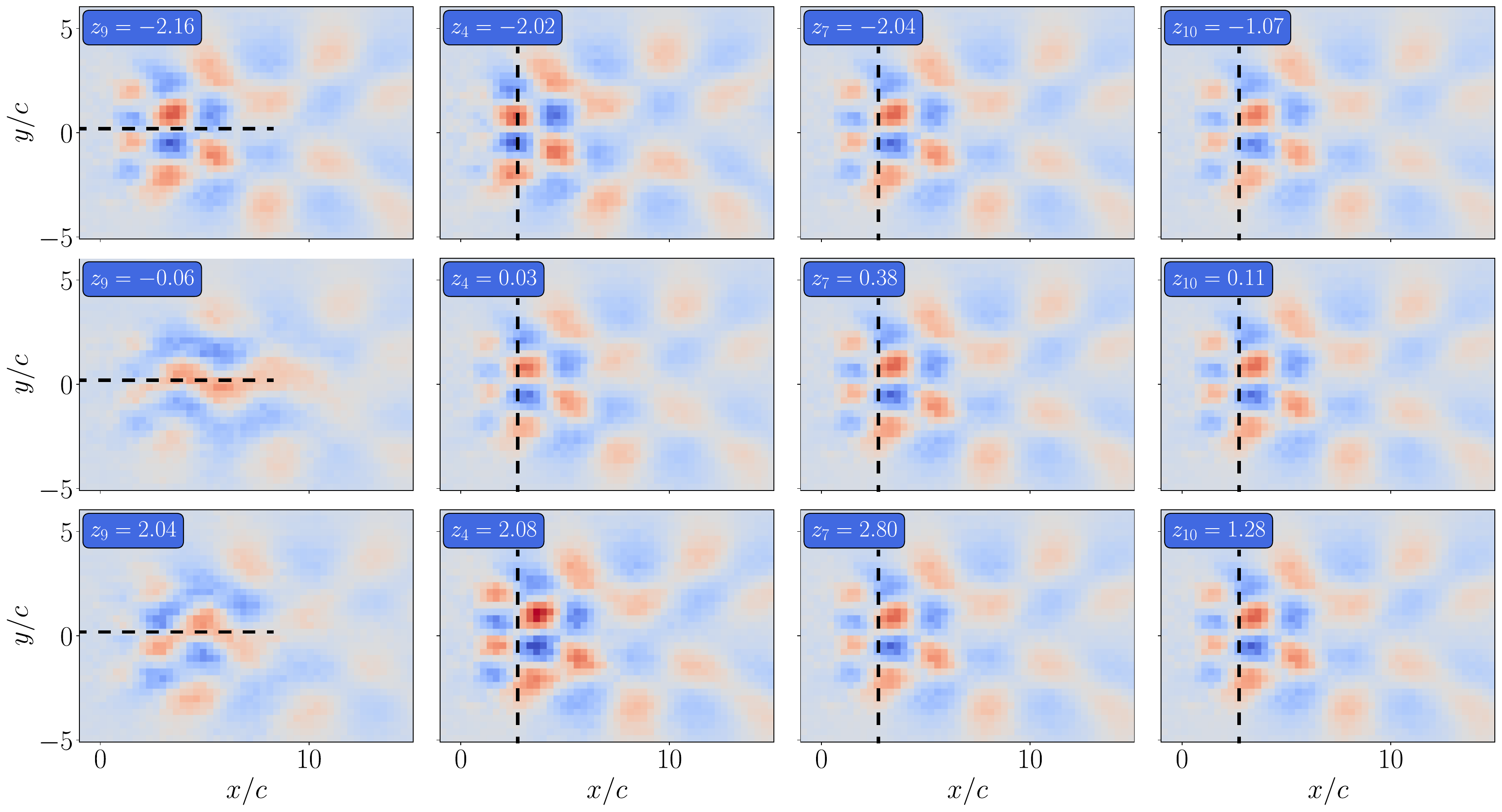}
    \caption{Impact of single latent variables on the $u$-velocity in the reference snapshot  of the verification case for the $\beta$-VAE using a latent space dimension of $m=10$ and $\beta=10^{-4}$.
    In each column, nine latent variables are fixed to the corresponding latent space representation of the snapshot and one variable takes equidistant values ranging from the minimum to the maximum value obtained
    for the validation set.
    Color code is the same for all examples in this figure, ranging from -1.67 (dark blue) to 1.94 (dark red).}
    \label{fig:Case1_VAE_Latent10_modes}
\end{figure}

\smallskip
To verify how the modes for the active and inactive latent variables differ, we  generate modes for the four most active latent variables of the latent space dimension $m=10$ in combination with the intermediate weight values, i.e., the red curves in Fig.~\ref{fig:case1_std}. The standard deviation is used for identification in combination with the UAE and OAE models. For the $\beta$-VAE ($\beta=10^{-4}$) we employ the three latent variables that display the largest standard deviation ($z_4, z_7, z_9$) in Fig.~\ref{fig:case1_std} together with $z_{10}$ that displays a large KL divergence in Fig. \ref{fig:case1_kld}.  
To safe space, we restrict ourselves to the influence of the modes on the horizontal $u$-velocity fields in a confined spatial domain as outlined in Figs.~\ref{fig:Case1_UAE_Latent10_modes}-~\ref{fig:Case1_VAE_Latent10_modes}. Consideration of the vertical velocity leads to the same conclusions.

Figure~\ref{fig:Case1_UAE_Latent10_modes} shows the UAE modes obtained for  $\nu=10^{-1}$, where default values for $m=10$ read {$z_1=-0.47$, $z_7=0.19$, $z_{10}=-0.10$ and $z_8=0.20$.}
The modes for the most active latent variable $z_1$ displayed in the first column clearly induce a shift in vertical direction. Similarly, the modes of the second most active variable $z_7$ displayed in the second column  introduce a horizontal shift. 
 The behavior is similar to that observed in the first columns of  Figs.~\ref{fig:case1_u_modes1} and~\ref{fig:case1_u_modes2}.
For the less active latent variables $z_{10}$ and $z_8$, the outputs are nearly identical for every latent value tested. 
This supports the assumption that the UAE model successfully identified the two crucial latent variables while being trained with a latent dimension of $m=10$. 
The inactiveness of latent variables in the UAE is in line with the findings by Kim et al.~\cite{KIM2021148}, in which the correlation penalty leads to inactive channels of encoded feature maps.

Figure~\ref{fig:Case1_VAE_Latent10_modes} displays the modes for the four most active latent variables of the $\beta$-VAE in combination with $\beta=10^{-4}$, where default values for $m=10$ are {$z_9=-1.90$, $z_4=0.89$, $z_{7}=2.14$ and $z_{10}=0.84$}. Again, the modes for the most active latent variables $z_9$ and $z_4$ display strong similarities to the modes investigated for $m=2$ in Figs.~\ref{fig:case1_u_modes1} and ~\ref{fig:case1_u_modes2}. 
Varying the value of the third most active latent variable $z_7$ does not significantly change the output of the decoder. Note that although $z_{10}$ has the lowest standard deviation, we also show the modes for $z_{10}$ because it has the second highest KL divergence. Interestingly, the output of the decoder also does not significantly alter in response to the value of $z_{10}$. Hence a large KL divergence without considering the standard deviation seems no clear indicator.

For the OAE we obtain slightly different results, as Fig.~\ref{fig:Case1_OAE_Latent10_modes} shows for the mode study based on $\lambda=10^{-4}$. Here default values for $m=10$ refer to {$z_3=-0.01$, $z_5=0.10$, $z_{1}=-0.03$ and $z_{10}=-0.03$}.
Horizontal shifts induced by the highest ranked latent variable $z_3$ and the third highest ranked latent variable $z_1$ are in close agreement. They are supplemented by vertical shifts introduced through changing the value of the second highest ranked variable $z_5$. The influence of $z_{10}$ on the encoder data is minor. Although the results are consistent with the data shown for $m=2$, the reduction to two modes is not so obvious in this case.

\subsubsection{Pruning}
In contrast to separating active from inactive latent variables during a post-processing step after the training, inactive latent variables could also be identified and pruned during the training process. 
The $i$th latent variable is pruned by setting the $i$th row of the weight matrix and the $i$th component of the bias of the final layer in the encoder to zero. By pruning inactive latent variables during training, these variables do not take part in the reconstruction of the input anymore and the models learn to only use the active latent variables. We tested such an approach for the UAE with a latent space dimension of $m=10$, in which an experiment-based threshold value of $0.07$ for the standard deviation of the normalized latent variables was employed to identify inactive latent variables, starting from the 500th epoch. A normalized threshold value slightly above $0.05$ can also be used to identify the active latent variables in a post-processing step.

The results were compared to a post-processing approach, where the two most active latent variables are identified after the training and the other latent variables are set to their mean, as for example suggested in~\cite{kang:2022}. 
We can confirm that the pruning approach successfully identified the two most active latent variables in several cases. 
However, no significant improvement of the reconstruction accuracy and the correlation coefficient was achieved compared to the post-processing approach. Further work on the pruning approach will be conducted in future research.

\subsection{Industrial Application: Aircraft Ditching Loads}
The second case serves to validate the use of the autoencoder models for more complex problems.  The particular application refers to the data-driven surrogate model of ditching load and deformation predictions. In a previous study~\cite{schwarz:2024}, different combinations of CAE and LSTM networks well as a Koopman autoencoder~\cite{lusch:2018} were assessed to predict the spatio-temporal evolution of ditching loads, with a \textit{convolutional encoder--LSTM--convolutional decoder} combination yielding the best results. In this paper, the focus lies on the disentanglement and analysis of the latent variables. 

The data set contains simulated ditching loads on the fuselage of a DLR-D150 aircraft, which is of similar size to an Airbus A320. The simulations were performed with the 3 degrees of freedom method \emph{ditch}~\cite{bensch:2003}, which considers horizontal, vertical and pitch motion of the aircraft in response to hydrodynamic loads, aerodynamic loads, inertia forces and thrust. Due to the resulting symmetry, only one half of the fuselage is used. A detailed description of the data generation and processing is given in~\cite{schwarz:2024}. 
The training set contains data from 323 simulations, which differ in the horizontal and vertical initial velocities. A validation set containing data from 20 simulations is used to compare the models for different weighting values of the disentanglement contributions to the loss functions.
For each simulation, around 18-35 time step images of size $128\times 128$, representing equidistantly spaced circumferential and longitudinal coordinates of the fuselage, are included in the data sets. The equidistant time step corresponds to 0.1 seconds.

\begin{figure}[h!]
    \centering
    \includegraphics[width=\linewidth]{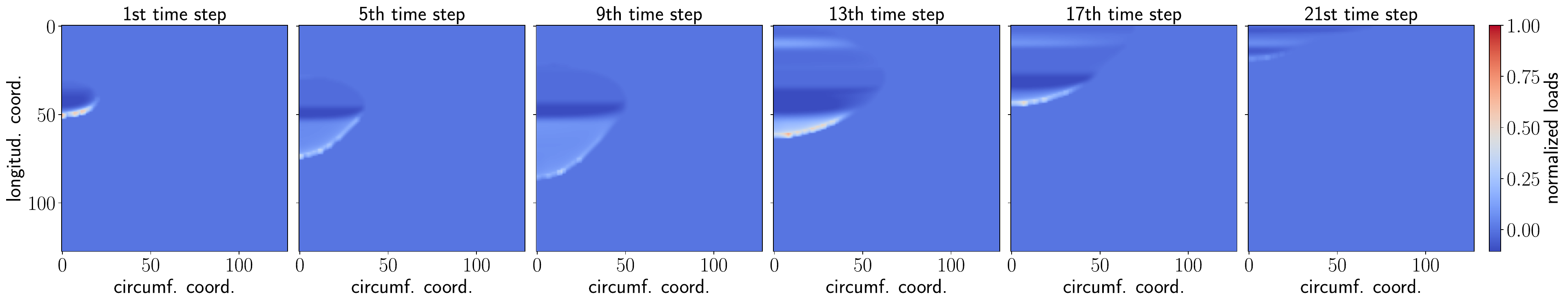}
    \caption{Exemplary spatio-temporal evolution of simulated ditching loads.}
    \label{fig:examplary_ditching_loads}
\end{figure}

The assessment of the autoencoder models is based on typical results for a temporal load series that were obtained from the trained models, see also \cite{schwarz:2024}. An exemplary temporal development of the hull loads is shown in Fig.~\ref{fig:examplary_ditching_loads} and shows the five typical phases. 
The first impact phase usually results in an initially  confined  circumferential load area (initial phase; 1st step), followed by a widening towards a larger longitudinal and circumferential load area (early phase; 5th, 9th step). For the subsequent phases, positive load signatures are usually slender crescent-shaped and limited to a smaller spatial area (late phase; 13th, 17th step). Moreover, they are followed by a more pronounced under pressure region in downstream direction. The final loads of rear fuselage are often small in amplitude and area (final phase; 21st step).

\subsubsection{Autoencoder Configuration}
The structure of the encoder and the decoder is based on the previous study~\cite{schwarz:2024} and outlined in  Table~\ref{tab:cae_ditch}. The baseline study employs a latent space dimension of $m=10$, which lead to satisfactory results in the previous research~\cite{schwarz:2024}, while keeping the latent space compact to allow for future research on interpretability. 
\begin{table}[h!]
\caption{Structure of encoder and decoder. The kernel size is $3\times 3$, and the stride is 2 for all Conv2d and ConvTranspose2d layers. For the $\beta$-VAE, two Linear(10) layers representing $\boldsymbol{\mu}$ and $\log (\boldsymbol{\sigma}^2)$ and the calculation of $\mathbf{z} = \boldsymbol{\mu} + \boldsymbol{\sigma} \odot \boldsymbol{\epsilon}$ replace the single Linear(10) layer in the encoder.}
\centering
\begin{tabular}{cccc}
    \hline
    Encoder & Output shape & Decoder & Output shape\\
    \hline
 Input & (3,128,128,1) & Linear(4096) & (4096)\\ 
     Conv2d(out\_channels=8) & (64,64,8) & Unflatten(8,8,64) & (8,8,64)\\
     Conv2d(out\_channels=16) & (32,32,16) & ConvTranspose2d(out\_channels=32) & (16,16,32)\\
     Conv2d(out\_channels=32) & (16,16,32) & ConvTranspose2d(out\_channels=16) & (32,32,16) \\
     Conv2d(out\_channels=64) & (8,8,64) &  ConvTranspose2d(out\_channels=8) & (64,64,8) \\
     Flatten() & (4096)  & ConvTranspose2d(out\_channels=1) & (128,128,1)\\  
     Linear(10) & (10)  \\
    \hline
\end{tabular}
\label{tab:cae_ditch}
\end{table}
A \textsf{LeakyReLU}~\cite{maas:2013} activation function  with slope parameter $\alpha=0.01$ was employed for the hidden layers in the encoder and the decoder. The output layers of the encoder and the decoder are linear. 
Models are trained for 500 epochs using Adam with the default learning rate of $1\cdot10^{-3}$ and a minibatch size of 128.

\subsubsection{Influence of Loss Function Weights}
Similar to the verification study in Sec. \ref{sec:periodicflow_loss_function_weights} we compare the reconstruction losses on the validation set and use the determinant of the correlation matrix $\det(\mathbf{R})$ to simultaneously  measure the disentanglement of the latent variables for different weights $\beta$, $\lambda$ and $\nu$ of the disentanglement contribution to the loss
function.  Again, both contributions to the loss function are used in every epoch.  To obtain some guidance on judging the results for the different $\beta$, $\lambda$ and $\nu$ values, we additionally trained a standard convolutional autoencoder without any disentanglement constraints in the latent space, and obtained $\det(\mathbf{R})=0.014$ and a reconstruction error of $2.53\cdot10^{-6}$. 

\begin{figure}[h!]
    \centering
    \includegraphics[width=0.8\textwidth]{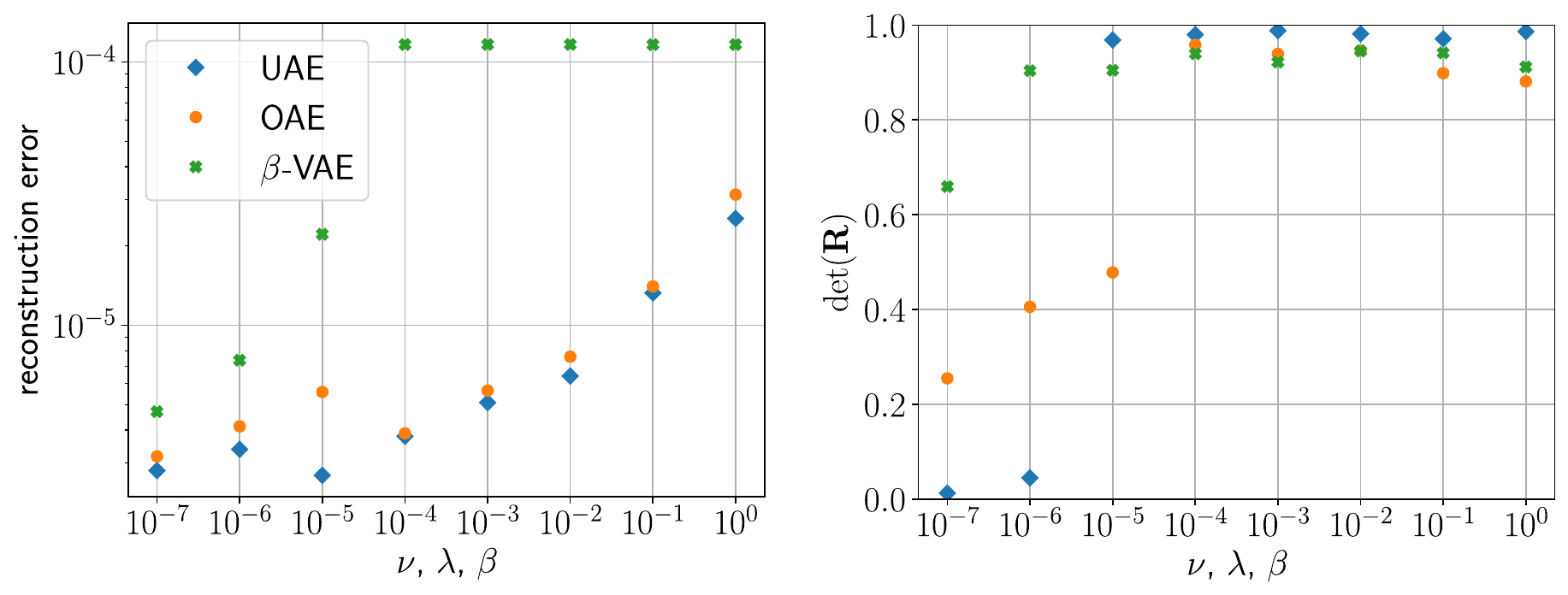}
    \caption{Reconstruction errors on the validation set and resulting values of $\det(\mathbf{R})$ obtained by the different models for the ditching application using $m=10$ latent variables. 
    }
    \label{fig:D150_loss_factors_vs_rec_losses_detR}
\end{figure}

Figure~\ref{fig:D150_loss_factors_vs_rec_losses_detR} depicts the influence of the weights on the reconstruction error (left) and the determinant of the correlation matrix (right).
The two deterministic models again reveal a robust behavior. For the investigated range of values for $\nu$ and $\lambda$, the minimization of the reconstruction loss does not get stuck in the beginning, although higher values 
usually yield higher reconstruction errors. The UAE, represented by the blue symbols, achieves the lowest reconstruction error and returns $\det(\mathbf{R}) \approx 1$ except for $\nu=10^{-6}$ and $\nu=10^{-7}$, where $\det(\mathbf{R})$ is close to zero, suggesting that $\nu\le 10^{-6}$ is too small to enforce uncorrelated latent variables. For the OAE, marked in orange, $\det(\mathbf{R})$ is mostly close to 0.9 and smaller than $\det(\mathbf{R})$ for the UAE when $\lambda\ge10^{-4}$. Although  $\det(\mathbf{R})$ values associated with $\lambda\le10^{-6}$ are higher for the OAE than for the corresponding UAE, the determinant only reaches values about 0.4 and 0.25. These values are quite small, indicating an undesirable correlation level between the latent variables.
On the contrary, one can observe that the $\beta$-VAE represented by the green symbols, only minimizes the reconstruction loss when $\beta$ is set smaller than $\beta=10^{-5}$, 
 which is the largest value, for which the minimization of the reconstruction loss does not get stuck in the beginning. For $\beta=10^{-4}$, we also tested a warm-up phase, in which $\beta$ gradually increased to $10^{-4}$ at the beginning of training. Without going into detail we note that, although the reconstruction loss is minimized at first, it eventually increases to around $1.2\cdot10^{-4}$ again, corresponding to the same reconstruction error level as for the greater values for $\beta$.
In all cases, the reconstruction error for the $\beta$-VAE exceeds the reconstruction error of the deterministic models. 
The values for $\det(\mathbf{R})$ are greater than $0.9$ for $\beta=10^{-5}$ and $\beta=10^{-6}$. For $\beta=10^{-7}$, it is below $0.7$, indicating a small interval of suitable $\beta$-values that achieve both, a disentanglement of the latent variables and a satisfactory reconstruction capability.

\subsubsection{Mode Analysis}
\label{sec:modean_D150}
\begin{figure}[h!]
    \centering
    \includegraphics[width=\textwidth]{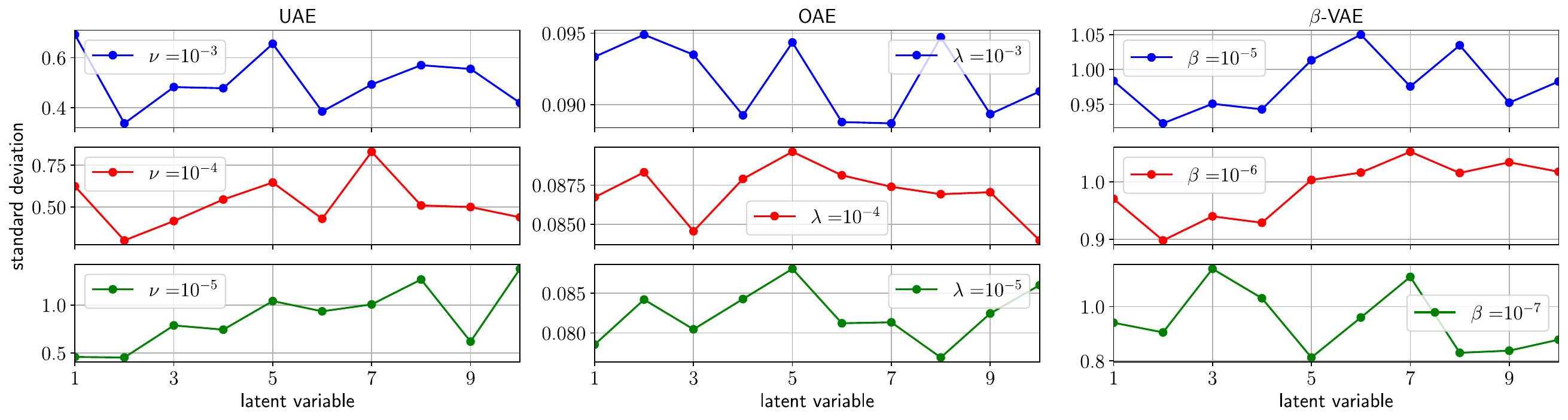}
    \caption{Standard deviations of the latent variables 
    obtained in the application case for a latent space dimension of $m=10$. Results for three different weights of the disentanglement contribution to the loss function are shown corresponding to Figure \ref{fig:D150_loss_factors_vs_rec_losses_detR}.}
    \label{fig:D150_Latent10_std}
\end{figure}

\begin{figure}[h!]
    \centering
    \includegraphics[width=\textwidth]{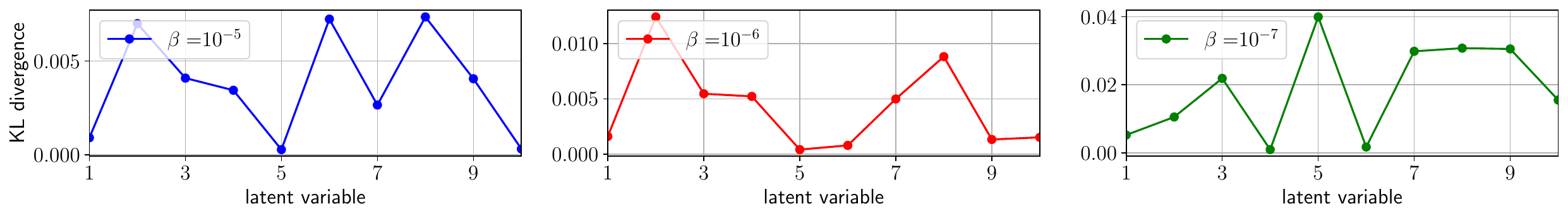}
    \caption{KL divergence of the latent variables for the $\beta$-VAE  obtained in the application case for a latent space dimension of $m=10$. Results for three different values of $\beta$ in the loss function are shown corresponding to Figure \ref{fig:D150_loss_factors_vs_rec_losses_detR}.}
    \label{fig:D150_Latent10_vae_KLD}
\end{figure}

Based on the studies from Sec. \ref{sec:modean-2Dm2}, Figs. \ref{fig:D150_Latent10_std} and \ref{fig:D150_Latent10_vae_KLD} depict the standard deviations of the latent variables of the three autoencoder models as well as the KL divergence of the $\beta$-VAE. The investigated three weight values for  $\nu$, $\lambda$ and $\beta$ were extracted from the results displayed in Fig. \ref{fig:D150_loss_factors_vs_rec_losses_detR}.

\smallskip
Considering the UAE on the left, we can observe that many latent variables appear to be active for all three investigated values of $\nu$.
For the large value $\nu=10^{-3}$ displayed in blue, which promises the best disentanglement whilst still delivering a fair amount of reconstruction accuracy, we can structure the latent space into two higher-ranked contributions $z_1, z_5$ and two second-ranked contributions
$z_8, z_9$.
As regards the OAE, the situation is similar. Due to the results shown in Fig. \ref{fig:D150_loss_factors_vs_rec_losses_detR}, we discard the lower values and focus on the largest meaningful  value $\lambda=10^{-3}$ displayed in blue.
 The related latent space can also be grouped into a dominating mode $z_2$ and two second-ranked modes $z_8, z_5$ followed by $z_3$.
For the $\beta$-VAE, Fig.
\ref{fig:D150_loss_factors_vs_rec_losses_detR} suggests to employ $\beta=10^{-6}$ as a baseline value, where the four most important latent variables refer to $z_7$ (top rank) and  $z_9$ slightly ahead of $z_6, z_8, z_{10}$, cf.  red curves in Fig. \ref{fig:D150_Latent10_std}. The KL divergence depicted in Fig.~\ref{fig:D150_Latent10_vae_KLD} suggests to assign $z_{6}$ to a lower priority. An interesting phenomenon occurs when comparing the standard deviation in Fig. \ref{fig:D150_Latent10_std} with the
KL divergence in Fig. \ref{fig:D150_Latent10_vae_KLD}. For the chosen  value $\beta=10^{-6}$, $z_2$ clearly shows the largest KL divergence, but also by far the smallest standard deviation. It turns out, that this latent variable is inactive with respect to the decoder output. This again underlines that a high KL divergence is only a good indicator of an active variable in combination with a sufficiently large standard deviation.

For each autoencoder model, we employ the four most active latent variables
to assess the associated modes. In contrast to the 
periodic flow example, we set all other latent variables to zero during the assessment, except the one we are analyzing. The reason is that the ditching data involves more different states and we believe that the choice of a suitable baseline state is not obvious. This also means that the studies do not consider a reference snapshot, since all but one latent variable are inactive.

\begin{figure}[h!]
    \centering
    \includegraphics[width=0.6\textwidth]{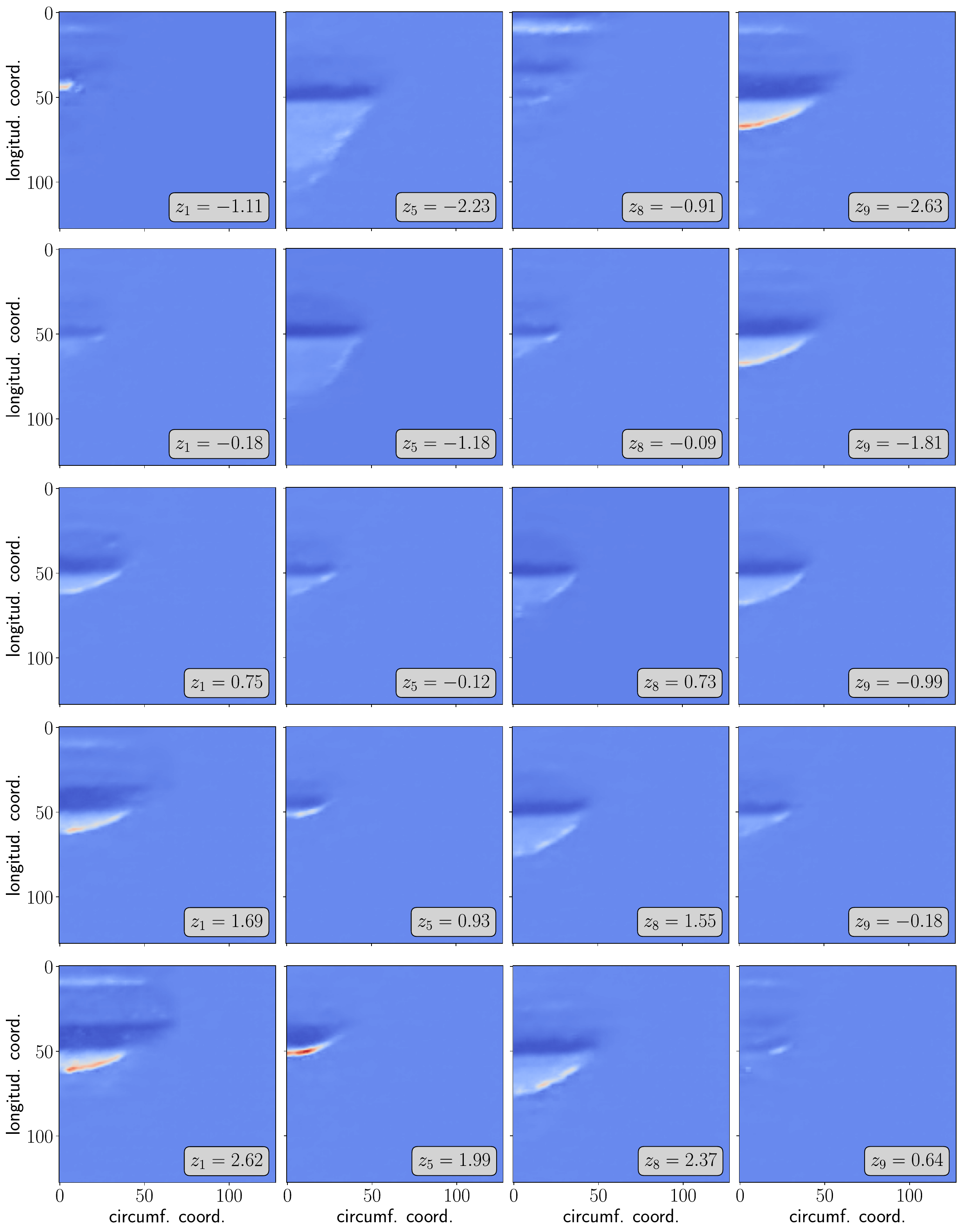}
    \caption{Impact of four latent variables with the highest standard deviation on the loads obtained from the UAE for the application case using a latent space dimension of $m=10$ and $\nu=10^{-3}$. 
    In each column,  one latent variable takes equidistant values ranging from the minimum to the maximum obtained on that variable on the validation set while the other latent variables are set to zero.
    Color code is the same for all examples in this figure, ranging from lowest (dark blue) to highest (dark red) values.}
    \label{fig:D150_UAE_Latent10_modes}
\end{figure}

\begin{figure}[h!]
    \centering
    \includegraphics[width=0.6\textwidth]{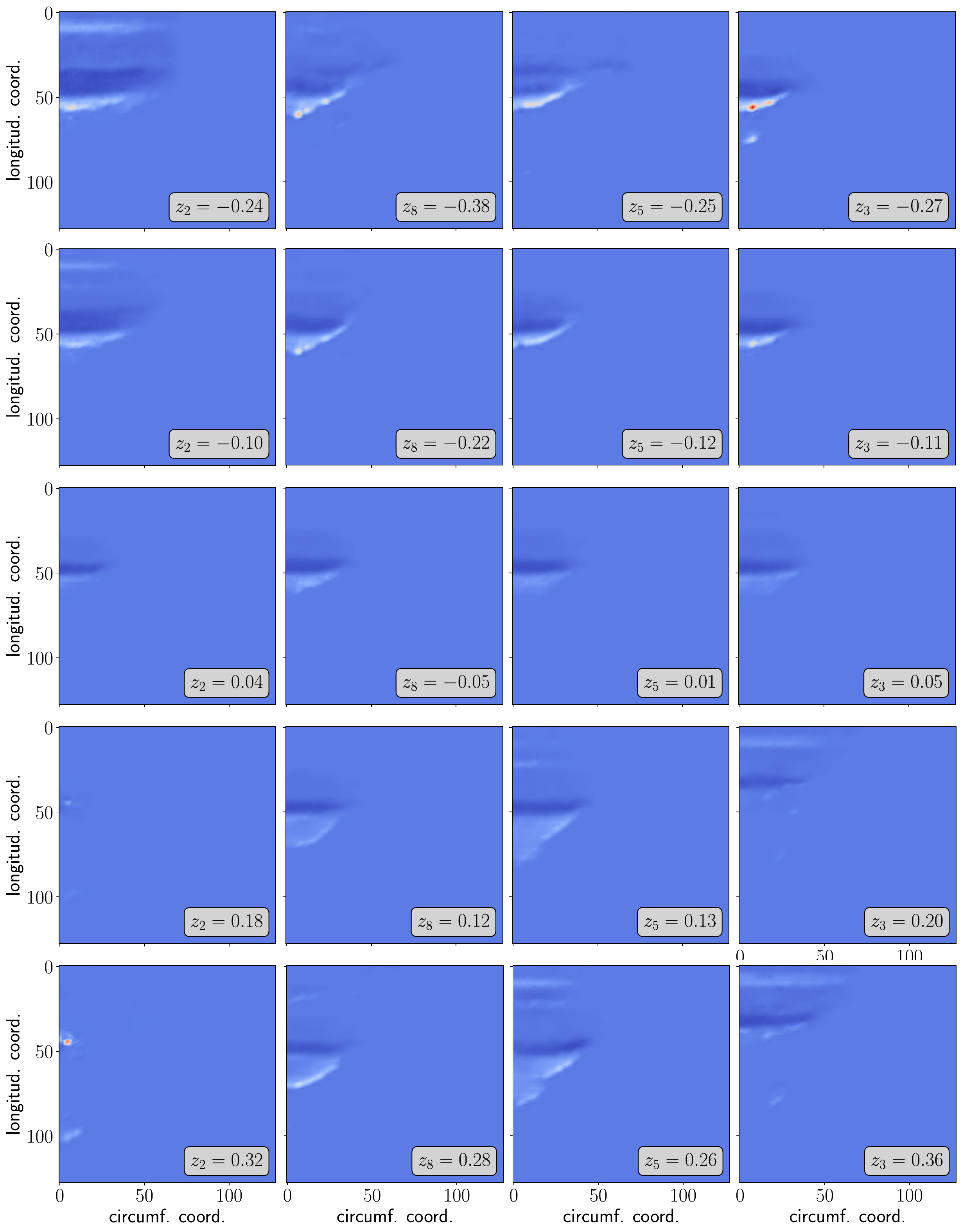}
    \caption{Impact of the four latent variables with the highest standard deviation on the loads obtained from the OAE for the application case using a latent space dimension of $m=10$ and $\lambda=10^{-3}$. 
    In each column, one latent variable takes equidistant values ranging from the minimum to the maximum obtained on that variable on the validation set while the other latent variables are set to zero.
    Color code is the same for all examples in this figure, ranging from lowest (dark blue) to highest (dark red) values.}
    \label{fig:D150_OAE_Latent10_modes}
\end{figure}

\begin{figure}[h!]
    \centering
    \includegraphics[width=0.6\textwidth]{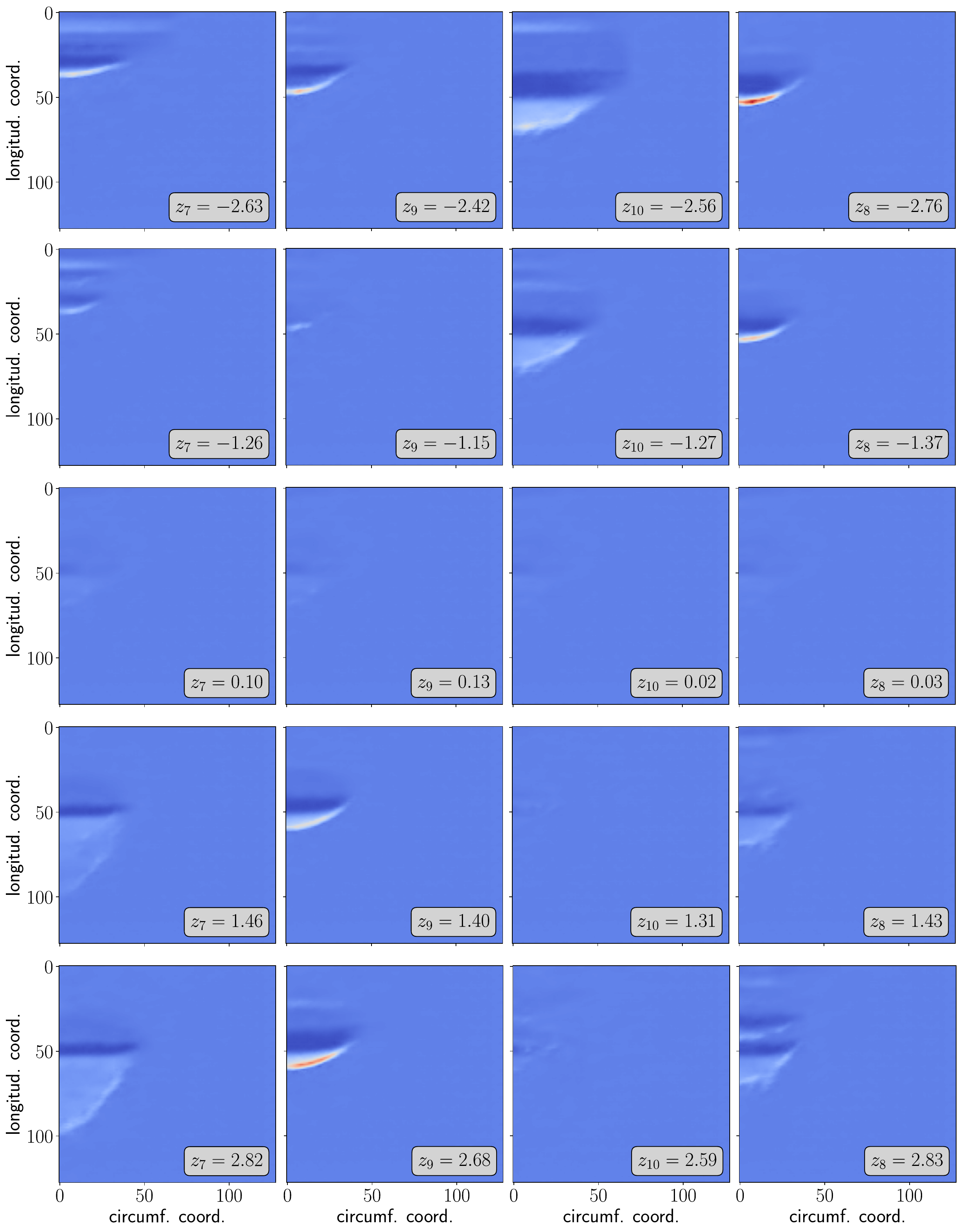}
    \caption{Impact of the four latent variables with the highest standard deviation on the loads obtained from the $\beta$-VAE for the application case using a latent space dimension of $m=10$ and $\beta=10^{-6}$. 
    In each column, one latent variable takes equidistant values ranging from the minimum to the maximum obtained on that variable on the validation set while the other latent variables are set to zero.
    Color code is the same for all examples in this figure, ranging from lowest (dark blue) to highest (dark red) values.}
    \label{fig:D150_VAE_Latent10_modes}
\end{figure}

Figure~\ref{fig:D150_UAE_Latent10_modes} illustrates the response of the load to the most 
active latent variables for the UAE. 
Considering the most active variable $z_1$, we can see that negative $z_1$-values only have a small impact on the output. Positive values seem to be related to the late phase of the impact, cf. steps 13 and 17 in Fig. \ref{fig:examplary_ditching_loads}, featuring a typical slender crescent-type  pressure load followed by a more pronounced suction load regime.
The second most active variable $z_5$ is contributing to the initial impact phase for positive values, cf. step 1 in Fig. \ref{fig:examplary_ditching_loads}, with a confined small area being exposed to high pressure loads. Negative $z_5$-values resemble the subsequent early phase, cf. steps 5 and 9 in Fig. \ref{fig:examplary_ditching_loads}, in which a large area of the fuselage is exposed to smaller pressure loads. 
Negative values in $z_8$ are associated to the final phase of the impact, in which  loads occur at the rear of the fuselage and begin to vanish, cf. step 21 in Fig. \ref{fig:examplary_ditching_loads}. For positive values, $z_8$ is contributing to the early impact phase, cf. steps 5 and 9 in Fig. \ref{fig:examplary_ditching_loads}. Small values of $z_9$ are related to the late phase and larger values to final phase of the ditching, cf. steps 13, 17 and 21 in Fig. \ref{fig:examplary_ditching_loads}.

The outputs generated by an investigation of the four most active latent variables obtained from the OAE are shown in Fig.~\ref{fig:D150_OAE_Latent10_modes}. The OAE results are generally more noisy than the corresponding UAE results. 
Although it is evident that certain latent variables contribute to the loads occurring in a particular phase of the ditching event,
for example, the spatial expansion of the load regime in the early impact phase captured by positive values of $z_5, z_8$ and a footprint of final phase loads represented by negative values of $z_2$, in most cases the patterns are not so clearly linked to a loading situation.

Looking at the modes associated with the four most active variables of the $\beta$-VAE  in Fig.~\ref{fig:D150_VAE_Latent10_modes}, we can see more distinct links with realistic load pattern again. The most active variable $z_7$ is associated to the final phase of the impact for negative values and to the early phase for positive values. Typical load pattern for the late phase resemble the structures reported by $z_9, z_8$, whereas $z_{10}$ supplements the load widening in circumferential direction.  
It is interesting to note, that the fourth highest ranked variable $z_6$ does not yield any meaningful output, although the fifth highest ranked variable $z_8$ does. The standard deviations of these two latent variables are nearly the same, cf. Fig. \ref{fig:D150_Latent10_std}. However, the KL divergence for $z_8$ is the second highest of all latent variables and the one for $z_6$ is nearly zero. For all remaining latent variables including $z_2$, which has the highest KL divergence in combination with the smallest standard deviation, the outputs are similar to the ones for $z_6$ and, thus, nearly identical. That is in line with the findings in Sec. \ref{sec:case1:higher_latent_dimension}, in which changes in the latent variable with the lowest standard deviation and highest KL divergence returned nearly constant outputs.  

\begin{figure}[h!]
    \centering
    \includegraphics[width=0.8\textwidth]{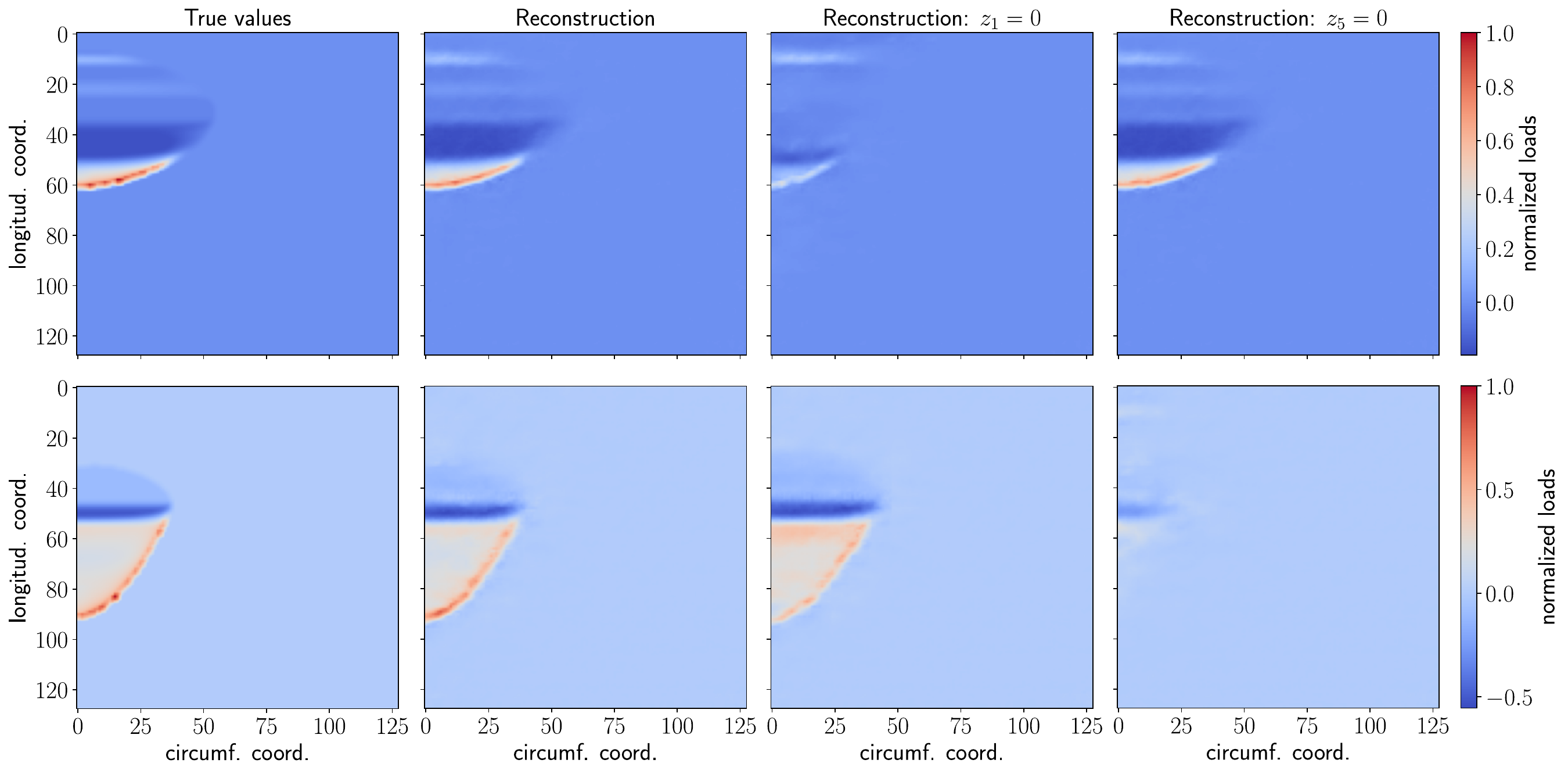}
    \caption{UAE reconstructions of the 16th (first row) and the 10th (second row) time step of a particular ditching case
    using a latent space dimension of $m=10$ and $\nu=10^{-3}$. 
    The suppressed variables in column three and four, i.e., $z_1$ and $z_5$, correspond to the most active and second most active latent variable, respectively.}
    \label{fig:D150_UAE_reconstructions}
\end{figure}

\smallskip
The resulting modes of the ditching loads are in general harder to analyze compared to the periodic flow case, as the temporal dynamics is more complex and over 300 different simulations with different input parameters were used to train the models. However, distinct influences of different active latent variables on different phases of a ditching event can be observed when the latent variables are disentangled.
One can verify that a latent variable is indeed associated to a specific impact phase by reconstructing a particular snapshot of this phase after setting this variable to 0. For this purpose, we reuse the second test case from~\cite{schwarz:2024}, which is simulated with the full-order model using a horizontal velocity of 80.75 m/s, a vertical velocity of 1.91 m/s and a pitch angle of $6^\circ$. 
As an example, we illustrate this for the two most active latent variables of the UAE, i.e., $z_1, z_5$, cf. Fig.~\ref{fig:D150_UAE_Latent10_modes}. Figure~\ref{fig:D150_UAE_reconstructions} shows snapshots for two particular time steps of the test case, their UAE-reconstruction using $m=10$ latent variables and the corresponding UAE-reconstruction obtained from setting either the most active latent variable $z_1$ or the second most active $z_5$ to zero. The selected time steps correspond to impact phases associated to the two most active latent variables, i.e., one from the early and one from the late phase, cf. Fig.~\ref{fig:D150_UAE_Latent10_modes}. 
Comparing the true values in the first column with the reconstructions using all $m=10$ latent variables shown in the second column displays a satisfactory agreement. If $z_1=0$ is set, the reconstruction of the 16th time step depicted in the first row is considered insufficient. However, if $z_5=0$ is set, the reconstruction is again very accurate.
A similar observation can be made for the 10th time step, which is shown in the second row. Removing the most active latent variable does not have a significant impact on the reconstruction quality. If $z_5=0$ is set, the reconstruction deteriorates significantly. 
It is evident that the corresponding latent variables have a significant influence on the quality of the reconstruction in the respective distinct phases. 

\subsubsection{Physics Awareness for Higher Latent Space Dimension}
Similar to Sec. \ref{sec:case1:higher_latent_dimension}, 
we train the models with a latent space dimension of 100 to see, how well the models work at finding a smaller number of active latent variables. Again, we use different weighting values for the contribution of the disentanglement part to the loss function, i.e., $\nu = \lambda = 10^{-3}, 10^{-4}, 10^{-5}$ as well as $\beta = 10^{-5}, 10^{-6}, 10^{-7}$ to analyze the impact of the loss term in the latent space.

\begin{figure}[h!]
    \centering
    \includegraphics[width=\textwidth]{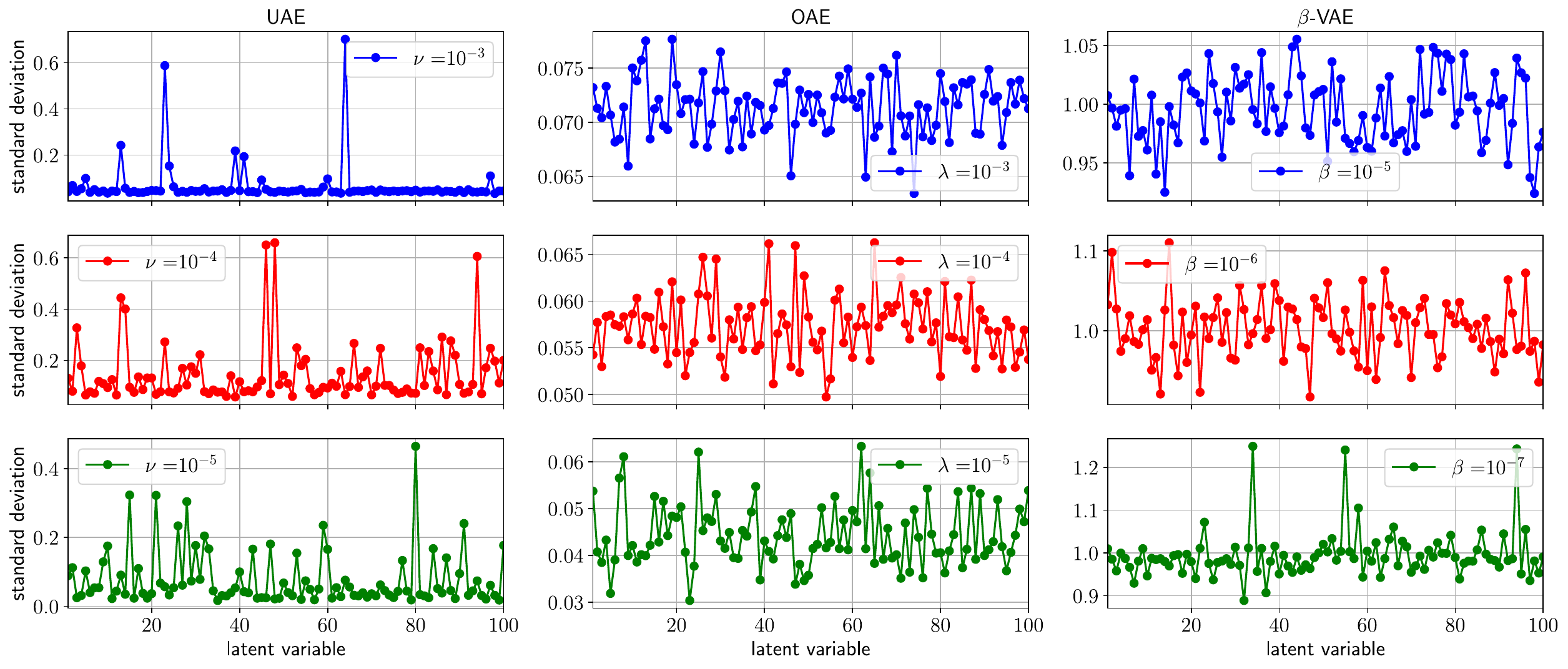}
    \caption{Standard deviations of the latent variables for all three autoencoder models using a latent space dimension of $m=100$. Each model is trained with three different weights of the disentanglement
contribution to the loss function.}
    \label{fig:D150_std}
\end{figure}

\begin{figure}[h!]
    \centering
    \includegraphics[width=\textwidth]{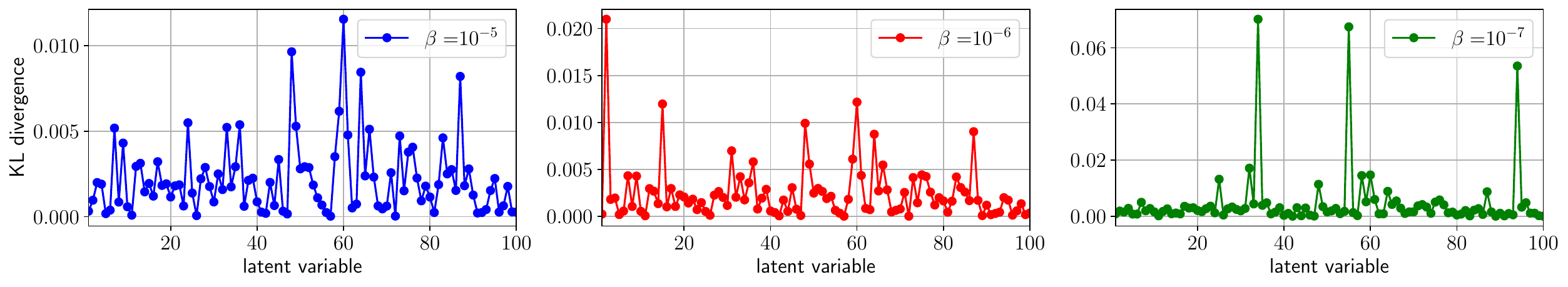}
    \caption{KL divergence for the latent variables of the $\beta$-VAE using a latent space dimension of $m=100$. The model is trained with
    three different values of $\beta$ in the loss function.}
    \label{fig:D150_Latent100_vae_KLD}
\end{figure}

Figure~\ref{fig:D150_std} shows the standard deviations of the different latent variables for all three autoencoder models using the different weighting values. We can observe a similar behavior as for the periodic flow example in Fig.~\ref{fig:case1_std}. Looking at the UAE, we can identify two distinct active latent variables in combination with the most pronounced disentanglement $\nu=10^{-3}$. Four more latent variables have a noticeably higher standard deviation than the remaining ones, from which four further variables stand out. By reducing the weight to $\nu=10^{-4}$ and $\nu=10^{-5}$, the number of active latent variables increases significantly. 
Observing the behavior of the OAE model shown in the second column, the standard deviations of the latent variables are again generally much closer to each other for $\lambda=10^{-3}$. For $\lambda=10^{-4}$ and $\lambda=10^{-5}$, the differences between the standard deviations become higher, but are never  as pronounced as for the UAE. 
Considering the $\beta$-VAE in the right column, we can observe that the standard deviations are also fairly close to each other for $\beta=10^{-5}$ and $\beta=10^{-6}$. For $\beta=10^{-7}$, there are three distinct latent variables with a clearly higher standard deviation.  
The corresponding KL divergences are depicted in Fig.~\ref{fig:D150_Latent100_vae_KLD}. As expected, higher KL divergences are returned for lower values of $\beta$. For $\beta=10^{-7}$, the three highest values are obtained for the three latent variables with the highest standard deviation in Fig \ref{fig:D150_std}.

\begin{figure}[h!]
    \centering
    \includegraphics[width=0.9\textwidth]{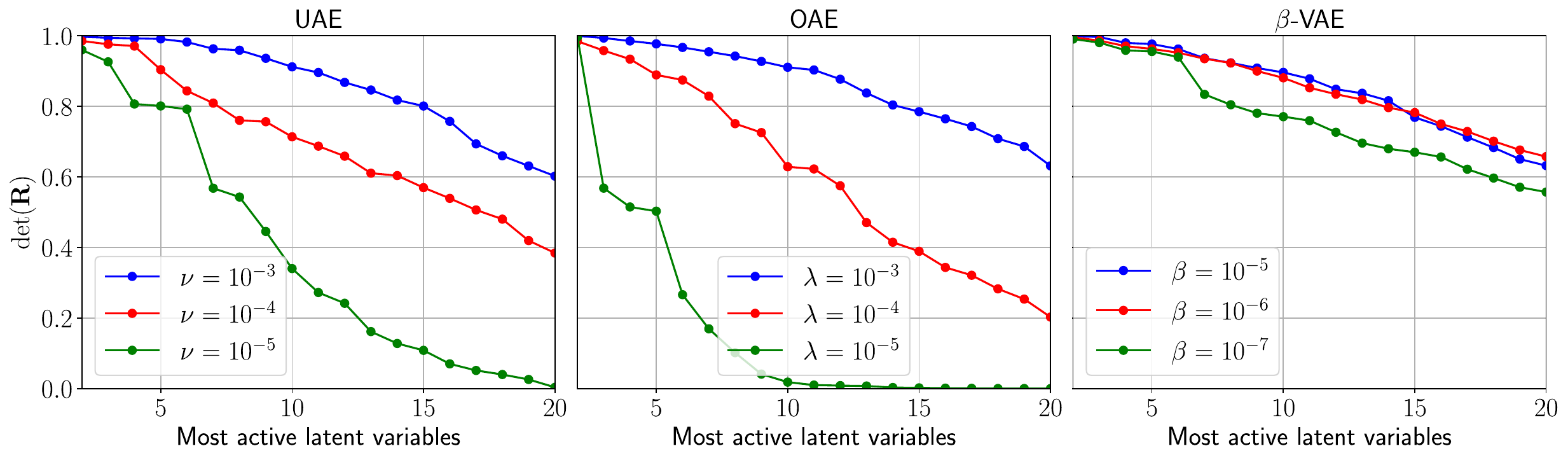}
    \caption{Values for $\det(\mathbf{R})$ using different amounts of the most active latent variables.}
    \label{fig:D150_UAE_Latent100_detR}
\end{figure}

Regarding $\beta$-VAE, Kang et al.~\cite{kang:2022} noted that considering $\det(\mathbf{R})$ using all latent variables is not sensible when the latent space is sparse, and suggested to consider combinations of a smaller number of latent variables. 
Similarly, we compute $\det(\mathbf{R})$ for the twenty most active latent variables to verify their independence. The corresponding results are illustrated in Fig.~\ref{fig:D150_UAE_Latent100_detR}. For all three models, an increased weight of the disentanglement contribution to the loss function yields higher values of $\det(\mathbf{R})$. The only exception is the $\beta$-VAE, for which the differences between $\beta=10^{-5}$ and $\beta=10^{-6}$ are negligible. Analyzing the UAE behavior for $\nu=10^{-3}$, one finds that the five most active latent variables are all approximately at $\det(\mathbf{R})\approx 1$, and $\det(\mathbf{R})$ decreases only slightly when using additional active variables.
For the 10 most active latent variables, $\det(\mathbf{R})$ is still greater than $0.9$. When reducing $\nu$, the value of $\det(\mathbf{R})$ decreases more rapidly, starting with five ($\nu=10^{-4}$) or even three ($\nu=10^{-5}$) variables.
The OAE gives similar results to the UAE for $\lambda=10^{-3}$. Using $\lambda=10^{-4}$ and $\lambda=10^{-5}$ induces a more rapid decrease in $\det(\mathbf{R})$ when increasing the amount of latent variables. 
For the $\beta$-VAE, $\beta=10^{-5}$ and $\beta=10^{-6}$ return similar values of $\det(\mathbf{R})$ as $\nu=\lambda=10^{-3}$ for the UAE and OAE, respectively, although the five most active latent variables yield slightly smaller values than the five most active for the UAE. From the seventh ranked variable, $\det(\mathbf{R})$ deviates from the higher values of $\beta$ for $\beta=10^{-7}$ and takes values smaller than $\approx0.83$.

\begin{figure}[h!]
    \centering
    \includegraphics[width=0.6\textwidth]{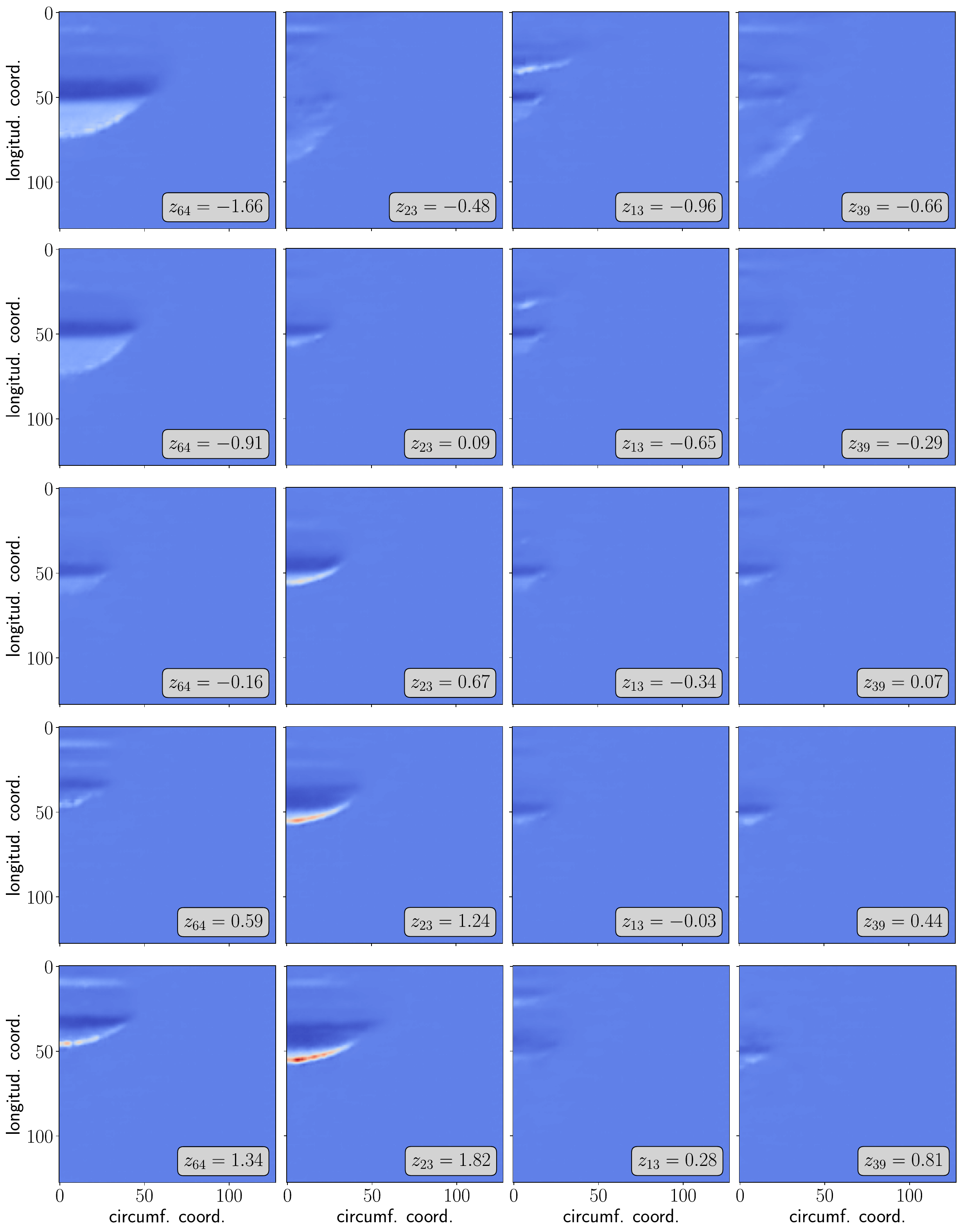}
    \caption{Impact of four latent variables with the highest standard deviation on the loads obtained from the UAE trained with a latent space dimension of $m=100$. In each column,  one latent variable takes equidistant values ranging from the minimum to the maximum obtained on that variable on the validation set while the other latent variables are set to zero.
    Color code is the same for all examples in this figure, ranging from lowest (dark blue) to highest (dark red) values.}
    \label{fig:D150_UAE_Latent100_modes}
\end{figure}

In Fig.~\ref{fig:D150_UAE_Latent100_modes}, the modes for the four most active latent variables of the UAE using $\nu=10^{-3}$ are displayed in combination with $m=100$. The modes are generated exactly as described in Sec. \ref{sec:modean_D150}. For the two most active latent variables $z_{64}$ and $z_{23}$ results agree with the modes associated with $z_1$ and $z_5$ in the $m=10$ configuration, cf. Fig.~\ref{fig:D150_UAE_Latent10_modes}. The main differences are the flipped order of the first column, which is again related to the late phase,  as well as differences in the intensities of the loads.  The two following latent variables of the $m=100$ configuration $z_{13}$ and $z_{39}$, however, have a smaller impact on the output. This could also be expected from the different standard deviations depicted in Fig. \ref{fig:D150_std}.
Without going into detail, we report that the $\beta$-VAE also returned reasonable leading modes for $\beta=10^{-7}$ and $m=100$.

\section{Conclusion}
\label{sec:4}
The paper compares the performance of two deterministic autoencoder models and the $\beta$-variational autoencoder 
for the dimension reduction of two-mode periodic benchmark flow data and multi-mode aircraft ditching load data.  Emphasis is placed on the disentanglement of the latent variables and the analysis of the resulting modes. Considering the reconstruction accuracy and the level of disentanglement in the latent space, the uncorrelated autoencoder (UAE) outperformed the $\beta$-VAE, while being deterministic and easier to train with respect to the choice of the hyperparameter that balances the reconstruction loss and latent space loss in both test cases.
Therefore, if only reconstruction accuracy and disentanglement of latent variables are desired in a surrogate model, but not specifically a normally distributed latent space, the UAE may more easily provide satisfactory results.
Similarly to $\beta$-VAE, the UAE can identify a small and limited number of truly active latent variables when the model is trained with a larger latent space dimension than required. Such active latent variables can be easily identified by their standard deviation. 
The second deterministic autoencoder model tested, the orthogonal autoencoder (OAE), was also able to return small correlation coefficients between the latent variables. However, it was not able to identify the active latent variables in this study when trained with a higher latent space dimension.
The analysis of the ditching case provided a better understanding of how the different latent variables contribute to the reconstruction. It was observed that different latent variables concentrate at different temporal phases of the impact. This can be of importance in future work, when deformation-induced load changes are to be included in the model.

\section*{Author Contributions}
\textbf{Henning Schwarz}: Conceptualization, Writing – original draft, Writing – review \& editing, Data curation, Formal analysis, Methodology, Software, Investigation, Validation, Visualization.
\textbf{Pyei Phyo Lin}: Writing - Review \& Editing, Conceptualization, Methodology.
\textbf{Jens-Peter M. Zemke}: Writing - Original Draft, Writing - Review \& Editing, Conceptualization, Methodology.
\textbf{Thomas Rung}: Funding acquisition, Writing – original draft, Writing – review \& editing, Conceptualization, Methodology, Project administration, Resources, Supervision.

\section*{Declaration of Interests}
The authors declare that they have no known competing financial interests or personal relationships that could have appeared to influence the work reported in this paper.

\section*{Data Availability}
The periodic flow dataset was created by Solera-Rico et al. \cite{solera-rico:2024} and is publicly available under \url{https://doi.org/10.5281/zenodo.10501216}~\cite{solera_rico_2024_dataset}.

\section*{Acknowledgements}
H.S., P.P.L., \ and T.R.\ acknowledge support by the German Federal Ministry for Economic Affairs and  Climate Action under aegis of the ''Luftfahrtforschungsprogramm LuFo VI'' project HYMNE (grant 20E2218A)  
and by the Clean Aviation project FASTER-H2, funded by the European Union under Grant Agreement No. 101101978. The views and opinions expressed are solely those of the author(s) and do not necessarily reflect those of the European Union or Clean Aviation. Neither the European Union nor Clean Aviation can be held responsible for them.
This paper is a contribution to the research training group RTG 2583 on ''Modeling, Simulation and Optimization of Fluid Dynamic Applications''  funded by the Deutsche Forschungsgemeinschaft (DFG). The authors acknowledge the support of TU Hamburg's Machine Learning in Engineering (MLE) initiative.

\bibliographystyle{elsarticle-num} 
\bibliography{mybibliography}
\end{document}